\pdfoutput=1
\documentclass{article}

    \PassOptionsToPackage{numbers, compress}{natbib}

    \usepackage[preprint]{neurips_2025}

\usepackage[utf8]{inputenc} 
\usepackage[T1]{fontenc}    
\usepackage{hyperref}       
\usepackage{url}            
\usepackage{booktabs}       
\usepackage{amsfonts}       
\usepackage{nicefrac}       
\usepackage{microtype}      
\usepackage{xcolor}
\usepackage{graphicx}
\graphicspath{{fig/main/}} 
\usepackage{float}
\usepackage{amsfonts}       
\usepackage{amsmath}        
\usepackage{multirow}  
\usepackage{algorithm}
\usepackage{algpseudocode}
\usepackage{caption}        
\usepackage{wrapfig}
\usepackage{enumitem}  
\usepackage{float} 
\usepackage{tabularx}
\usepackage{placeins}  
\usepackage{cleveref}
\crefname{figure}{Fig.}{Figs.}
\usepackage{subcaption}
\usepackage{siunitx}
\usepackage{boldline, multirow}
\NewExpandableDocumentCommand\mcc{O{1}m}{\multicolumn{#1}{c}{#2}}

\newcommand{\attnmap}{\mathbf{A}}
\newcommand{\paragrapht}[1]{\vspace{-6pt}\paragraph{#1}}

\usepackage[toc,page,header]{appendix}
\usepackage{minitoc}

\title{Where and How to Perturb: On the Design of Perturbation Guidance in Diffusion and Flow Models}

\author{
Donghoon Ahn$^{\bullet1}$ \qquad 
Jiwon Kang$^{\bullet1}$ \vspace{0.3em}\\
\textbf{Sanghyun Lee}$^{\circ1}$\qquad
\textbf{Minjae Kim}$^{\circ2}$ \qquad
\textbf{Jaewon Min}$^1$ \qquad
\textbf{Wooseok Jang}$^1$ \vspace{0.3em}\\
\textbf{Sangwu Lee}$^3$ \qquad
\textbf{Sayak Paul}$^4$ \qquad
\textbf{Susung Hong}$^5$ \qquad
\textbf{Seungryong Kim}\textsuperscript{$\dagger1$}
}

\begin{document}

\doparttoc
\faketableofcontents
\part{}

\maketitle

\begin{center}
    \vskip -2em
    {\normalsize
        $^1$KAIST AI \quad
        $^2$Korea University \quad
        $^3$Krea AI \quad
        $^4$Hugging Face \quad
        $^5$University of Washington
    \par}
    \vskip 1em
\end{center}

\begingroup
\renewcommand{\thefootnote}{}
\footnotetext{\scriptsize 
$\bullet$: Equally contributed as first author, 
$\circ$: Equally contributed as second author, 
\scriptsize $\dagger$: Corresponding author}
\endgroup

\begin{abstract}
Recent guidance methods in diffusion models steer reverse sampling by \textit{perturbing} the model to construct an implicit weak model and guide generation away from it. Among these approaches, attention perturbation has demonstrated strong empirical performance in unconditional scenarios where classifier-free guidance is not applicable. However, existing attention perturbation methods lack principled approaches for determining where perturbations should be applied, particularly in Diffusion Transformer (DiT) architectures where quality-relevant computations are distributed across layers. In this paper, we investigate the granularity of attention perturbations, ranging from the layer level down to individual attention heads, and discover that specific heads govern distinct visual concepts such as structure, style, and texture quality. Building on this insight, we propose ``HeadHunter", a systematic framework for iteratively selecting attention heads that align with user-centric objectives, enabling fine-grained control over generation quality and visual attributes. In addition, we introduce SoftPAG, which linearly interpolates each selected head’s attention map toward an identity matrix, providing a continuous knob to tune perturbation strength and suppress artifacts. Our approach not only mitigates the oversmoothing issues of existing layer-level perturbation but also enables targeted manipulation of specific visual styles through compositional head selection. We validate our method on modern large-scale DiT-based text-to-image models including Stable Diffusion 3 and FLUX.1, demonstrating superior performance in both general quality enhancement and style-specific guidance. Our work provides the first head-level analysis of attention perturbation in diffusion models, uncovering interpretable specialization within attention layers and enabling practical design of effective perturbation strategies. Our project page is available at: \href{https://cvlab-kaist.github.io/HeadHunter/}{https://cvlab-kaist.github.io/HeadHunter/}. 
\end{abstract}

\section{Introduction}

Diffusion models~\cite{ho2020ddpm, Song2020ddim, jascha2015, yang2019estgrad, yang2020sde, dhariwal2021diffusion, rombach2022high, dustin2023sdxl} and flow-matching models~\cite{Lipman2022FlowMatching, nanye2024sit, esser2024scaling} have gained great popularity in visual generation tasks, including images~\cite{rombach2022high, dustin2023sdxl, peebles2023scalable, chen2024pixart, esser2024scaling}, videos~\cite{ho2022video, Ho2022ImagenVH, Blattmann2023StableVD, Zhuoyi2025cogvideox}, 3D~\cite{poole2022dreamfusion,lin2023magic3d,wang2023prolificdreamer,shi2023mvdream}, and 4D content~\cite{Singer2023TextTo4D, Yu20244Real, Rundi2025cat4d}. The key behind the success of diffusion models lies in classifier-free guidance (CFG)~\cite{ho2022classifier}, which substantially enhances image generation quality during conditional inference. Despite its effectiveness, CFG has two key limitations. First, it applies only to conditional generation, limiting its applicability in unconditional settings such as inverse problems~\cite{zahra2021linear, Chung2022DiffusionPS, Song2023Pseudoinverse, Chung2022ImprovingDM}. Second, CFG often reduces sample diversity and leads to over-saturated or overly simplified outputs~\cite{, seyedmorteza2023cads, Kynkäänniemi2024Applying, Saharia2022PhotorealisticTD, tero2024}.

To address the limitations of CFG in unconditional scenarios, alternative guidance strategies~\cite{ahn2024self,hong2023improving,hong2024smoothed,tero2024,hyung2025spatiotemporal} have been proposed. These methods steer the denoising trajectory by perturbing the input or the model itself, or by training a weaker model, thereby guiding samples away from low-quality regions and toward the high-quality data manifold. Among these strategies, attention-layer perturbation approaches~\cite{ahn2024self,hong2024smoothed} continue to be a practical and widely explored strategy, as they can generate well-aligned weak models without additional training. Some papers explain these guidance methods from an energy perspective~\cite{hong2024smoothed} and draw connections to Hopfield networks~\cite{demircigil2017model, ramsauer2020hopfield}, which empirically work well with specific blocks (medium blocks of U-Net~\cite{ronneberger2015u} architecture) of attention. 

However, there is still limited understanding of where perturbations should be applied. In particular, Diffusion Transformer (DiT)~\cite{peebles2023scalable} architectures lack localized blocks responsible for global semantics, unlike U-Nets~\cite{ronneberger2015u}, and instead distribute this functionality more evenly across all layers~\cite{avrahami2024stable}. This structural difference makes it even more critical to carefully select perturbation targets to achieve effective guidance. 

To determine suitable perturbation targets, we first consider the underlying structure of DiT. In DiT, multi-head self-attention~\cite{vaswani2017attention} plays a central role in  modeling global dependencies. Each head attends to different aspects of the input, and prior work has shown that heads often specialize in distinct semantic features~\cite{vaswani2017attention,park2024cross,dosovitskiy2020image,gandelsman2023interpreting}. This suggests that attention heads, rather than entire layers, may serve as more precise and effective targets for perturbation.

Motivated by this, we explore finer-grained yet semantically meaningful computational units within attention layers: \textit{attention heads}. 
Interestingly, we observe that individual head-level perturbation guidance often capture interpretable visual concepts and specialize in tasks such as enhancing structural fidelity or injecting stylistic elements. Moreover, these functional roles can be composed by combining multiple heads. 

Building on these observations, we (i) analyze key properties of head-level perturbation guidance, including composability and controllability, (ii) propose \textbf{HeadHunter}, a systematic framework for retrieving heads that align with arbitrary objectives, and (iii) introduce \textbf{SoftPAG}, a variant of PAG~\cite{ahn2024self} that enables fine-grained, continuous control over guidance strength, mitigating over-smoothness and oversimplification caused by overly aggressive perturbations.

We demonstrate that HeadHunter not only outperforms layer-level perturbation guidance in general quality improvement but also enables targeted manipulation of visual styles, as supported by strong qualitative and quantitative results.

In summary, our main contributions are as follows:
\begin{itemize}[leftmargin=*, itemsep=2pt]
\item To the best of our knowledge, we are the first to apply perturbations at the level of {individual attention heads}, enabling fine-grained and concept-specific control.
\item We analyze the properties of head-level perturbation guidance in Diffusion Transformers (DiT), providing insights into the specialization and combination of such heads.
\item We propose {HeadHunter}, a systematic head selection framework for arbitrary objectives, and demonstrate its effectiveness in both general quality enhancement and style-specific guidance.
\item We introduce {SoftPAG}, which interpolates each selected head’s attention map toward the identity matrix, providing a continuous knob to tune perturbation strength and to mitigate oversaturation and oversimplification.
\end{itemize}

\section{Preliminaries}\label{sec:pre}


\paragrapht{Diffusion and flow matching models.}
Diffusion models~\cite{ho2020ddpm,sohl2015deep,song2020score} define a generative process as iterative denoising of a sample drawn from a Gaussian prior. The forward process gradually perturbs a clean data point $\mathbf{x}_0 \sim p_{\text{data}}$ through the noise schedule $(\alpha_t,\sigma_t)$, as
$\mathbf{x}_t = \alpha_t \mathbf{x}_0 + \sigma_t \boldsymbol{\epsilon}$,
where $\boldsymbol{\epsilon} \sim \mathcal{N}(\mathbf{0}, \mathbf{I})$ and $t \in [0,1]$. A neural network $\hat{\boldsymbol{\epsilon}}_\theta(\mathbf{x}_t,t)$ learns to predict the added noise $\boldsymbol{\epsilon}$, enabling reverse-time sampling from noise to data. On the other hand, flow matching~\cite{lipman2022flow,liu2022flow} provides a deterministic alternative by learning a continuous-time velocity field that guides a linear interpolation from noise to data:
$\mathbf{x}_t = (1 - t)\mathbf{x}_0 + t\,\boldsymbol{\epsilon}$, where $\boldsymbol{\epsilon} \sim \mathcal{N}(\mathbf{0}, \mathbf{I})$, and the associated velocity field is $\mathbf{u}(\mathbf{x}_t, t) = \boldsymbol{\epsilon} - \mathbf{x}_0$. A neural network $\hat{\mathbf{u}}_\theta(\mathbf{x}_t,t)$ is trained to approximate this target velocity.

\paragraph{Multi-head attention mechanism.}
Vaswani et al.~\citep{vaswani2017attention} introduced the multi-head attention mechanism, which projects queries, keys, and values into multiple subspaces and computes attention in parallel. At the $l$-th layer of the denoising network $\hat{\boldsymbol{\epsilon}}_\theta$, each head $h \in \{1, \dots, H_l\}$ applies distinct linear projections:
\begin{equation}
\mathbf{Q}_{l,h} = \mathbf{Q}_l \mathbf{W}_{l,h}^Q,\;
\mathbf{K}_{l,h} = \mathbf{K}_l \mathbf{W}_{l,h}^K,\;
\mathbf{V}_{l,h} = \mathbf{V}_l \mathbf{W}_{l,h}^V,
\end{equation}
where $\mathbf{Q}_l, \mathbf{K}_l, \mathbf{V}_l \in \mathbb{R}^{N \times d}$ are the query, key, and value matrices,
$\mathbf{W}_{l,h}^{Q,K,V} \in \mathbb{R}^{d \times \bar{d}}$ are the head-specific projections,
$\bar{d}$ is the per-head dimensionality, and $H_l$ is the number of attention heads in layer $l$.
The attention map and output for each head are
\begin{equation}
\label{eq:attention}
\mathbf{A}_{l,h} = \text{Softmax}\!\left(\frac{\mathbf{Q}_{l,h}\mathbf{K}_{l,h}^\top}{\sqrt{\bar{d}}}\right),
\quad
\mathbf{O}_{l,h} = \mathbf{A}_{l,h}\mathbf{V}_{l,h},
\end{equation}
and all head outputs are concatenated and projected:
\begin{equation}
\text{MultiHead}(\mathbf{Q}_l,\mathbf{K}_l,\mathbf{V}_l)
= \text{Concat}(\mathbf{O}_{l,1},\dots,\mathbf{O}_{l,H_l})\mathbf{W}_l^O,
\end{equation}
where $\mathbf{W}_l^O \in \mathbb{R}^{(H_l \bar{d}) \times d}$ is the output projection matrix.
Multi-head attention enables the model to capture diverse semantic relations across tokens~\cite{dosovitskiy2020image,park2024cross,gandelsman2023interpreting}.

\paragraph{Classifier-free guidance.}
Classifier-free guidance (CFG)~\cite{ho2022classifier} enhances conditional generation by extrapolating predictions from conditional and unconditional models:
\begin{equation}
\hat{\boldsymbol{\epsilon}}_{\text{CFG}} = (1 + w)\hat{\boldsymbol{\epsilon}}_{\text{cond}} - w\hat{\boldsymbol{\epsilon}}_{\text{uncond}},
\end{equation}
where $w$ is the guidance scale, and $\hat{\boldsymbol{\epsilon}}_{\text{cond}}$, $\hat{\boldsymbol{\epsilon}}_{\text{uncond}}$ denote the predicted noise with and without conditioning, respectively.

\paragraph{Attention perturbation guidance.}
\label{para:attention-perturbation-guidance}

As attention maps encode spatial and semantic structures~\cite{cao2023masactrl,nam2024dreammatcher}, perturbing them can guide the denoising trajectory away from suboptimal predictions, improving structural fidelity at test time~\cite{ahn2024self,hong2024smoothed,tero2024}. Analogous to classifier-free guidance~\cite{ho2022classifier}, attention perturbation guidance extrapolates between original and perturbed predictions:
\begin{equation}
\label{eq:attention-perturbation-guidance}
\hat{\boldsymbol{\epsilon}}_{\text{guided}} = (1 + w)\hat{\boldsymbol{\epsilon}}_{\text{original}} - w\hat{\boldsymbol{\epsilon}}_{\text{perturbed}},
\end{equation}
where $\hat{\boldsymbol{\epsilon}}_{\text{perturbed}}$ is obtained by modifying attention maps $\mathbf{A}_{l,h}$ in Eq.~\ref{eq:attention} during the forward pass.

Let $\mathcal{L}$ denote the perturbed layers and $H_l$ the number of heads in layer $l$. Perturbations are applied to $\mathbf{A}_{l,h}$ for all heads $h \in \{1, \dots, H_l\}$ and $l \in \mathcal{L}$. In perturbed-attention guidance (PAG)~\cite{ahn2024self}, the attention maps are replaced with the identity matrix, disabling contextual aggregation:
\[
\mathbf{A}_{l,h}^{(\text{PAG})} = \mathbf{I} \in \mathbb{R}^{N \times N}.
\]

\section{Motivation}
\label{sec:analysis}

\begin{figure}[h]
    \centering
    \includegraphics[width=\linewidth]{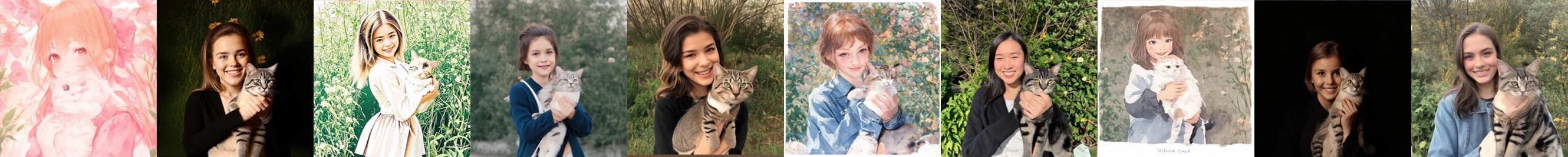}
    \caption{\textbf{Motivating example.} Each image is generated with PAG~\cite{ahn2024self}, where perturbation is applied to a single attention head within DiTs. Guiding with different perturbed \textit{attention heads} produces notably distinct results. Results for additional heads are provided in Appendix~\ref{sup:subsec:all-heads}. All images are generated with the prompt \textit{``smiling girl holding a cat, in a flower garden''} using \texttt{stable-diffusion-3-medium}. Each row corresponds to a single layer, with different heads perturbed across columns.}
    \label{fig:motivation}
\vspace{-10pt}
\end{figure}

Attention perturbation guidance~\cite{ahn2024self,hong2024smoothed,tero2024} steers generation away from weaker predictions by slightly altering the model’s forward pass. Thus, deciding what to weaken becomes critical, but principled methods for determining where in the diffusion network to apply perturbation remain underexplored. As a result, prior works often rely on heuristic selection of perturbation targets, such as the middle block of a U-Net~\cite{ronneberger2015u}, which is known to process high-level semantic information with global self-attention~\cite{hong2023improving,ahn2024self}.

However, unlike U-Net, transformer-based architectures as in DiTs~\cite{peebles2023scalable,chen2024pixart,esser2024scaling,blackforestlabs_flux1_dev,fal_aura_flow} lack a coarse-to-fine synthesis structure~\cite{avrahami2024stable}. This absence of an explicit bottleneck makes it challenging to directly identify layers responsible for high-level semantics. Instead, the model distributes semantic processing more uniformly across layers. In this setting, attention, particularly multi-head self-attention~\cite{vaswani2017attention}, plays a central role in modeling global dependencies without convolutions.

Each attention head processes different aspects of the input, and prior works in large language models and vision transformers have shown that heads specialize in capturing distinct semantic attributes~\cite{vaswani2017attention,park2024cross,dosovitskiy2020image,gandelsman2023interpreting}. This suggests that attention heads, rather than entire layers, may serve as more effective units for applying perturbation.

Motivated by this observation, we increase the granularity of attention perturbation from entire layers to individual \textit{attention heads}. In conventional approaches, referred to as \textbf{layer-level perturbation guidance} (or simply layer-level guidance), perturbations are applied uniformly to all attention heads within selected layers. In contrast, our proposed \textbf{head-level perturbation guidance} (or simply head-level guidance) selectively perturbs specific attention heads, enabling finer and more targeted control.
As shown in Fig.~\ref{fig:motivation}, perturbing different heads leads to clearly distinct effects in the generated images, indicating that heads act as \textit{semantically meaningful substructures}. This highlights the limitations of layer-level perturbation, which may be overly coarse and suboptimal. Even with carefully chosen layers, layer-level perturbation fails to leverage the modularity and functional diversity inherent in multi-head attention, motivating our focus on head-level guidance.




\begin{figure}[t]
  \centering
  \includegraphics[width=\columnwidth]{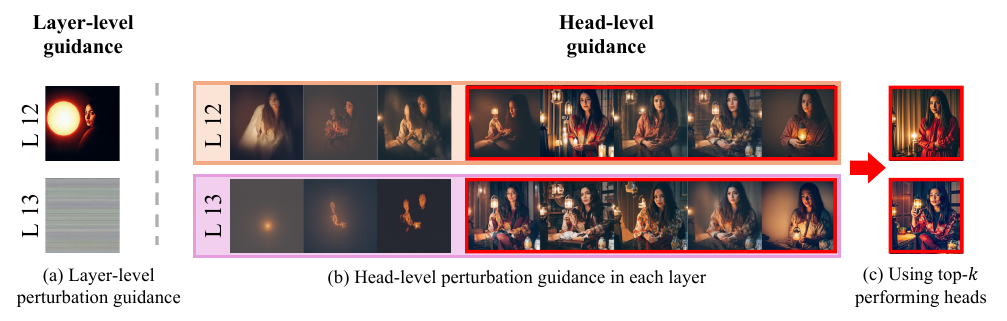}
  \caption{%
    \textbf{Generated images from head- and layer-level perturbation guidance.}
    (a) Results of layer-level perturbation guidance, where perturbation is applied to all heads in the layer.
    (b) Results of head-level perturbation guidance, where each result is obtained by independently applying perturbation to a single head of the layer. Red boxes indicate high-performing heads in terms of PickScore~\cite{kirstain2023pick}.
    (c) Perturbation guidance using only the high-performing heads identified in (b) yields higher-quality generations across both low-performing layers (L13) and well-performing ones (L12). The prompt \textit{“Turkish girl with lantern, dark room”} is used. }
      \label{fig:head-per-layer}
\end{figure}

\subsection{Analysis of head-level perturbation guidance}
\label{subsec:head-properties}

\paragraph{Definition.}
We define head-level perturbation guidance as applying perturbations selectively to a subset of \textit{attention heads} during the forward pass. Specifically, given a set $\mathcal{S} = \{(l_1, h_1), \dots, (l_m, h_m)\}$ of $m$ selected (layer, head) pairs, we replace their attention maps with identity matrices:
\begin{equation}
\label{eq:pag}
\mathbf{A}_{l,h}^{(\text{PAG})} = \mathbf{I} \quad \text{for } (l, h) \in \mathcal{S},
\end{equation}
where $\mathbf{A}_{l,h}^{(\text{PAG})}$ denotes the perturbed attention map of head $h$ at layer $l$, and $\mathbf{I}$ is the identity matrix used in PAG~\cite{ahn2024self}. While we illustrate this with PAG, the formulation also applies to other perturbation methods such as SEG~\cite{hong2024smoothed}, with each head perturbed independently. See Appendix~\ref{sup:head-perturbation-using-other-perturbations} for details.

\paragraph{Inefficiency of layer-level guidance due to intra-layer diversity.}

Even when we perturb the attention map of a single layer within diffusion models, it exhibits highly polysemantic attentive behavior, as the computation in individual heads occurs independently and diversely~\cite{elhage2021mathematical,park2024cross}. Therefore, applying the same perturbations across all heads in a layer fails to account for this intra-layer diversity, potentially disrupting the output from the guidance and leading to unintended side effects. As shown in Fig.~\ref{fig:head-per-layer}, some heads from supposedly poor layers (\texttt{L13} of Fig.~\ref{fig:head-per-layer} (a)) still produce high-quality results (Fig.~\ref{fig:head-per-layer} (b)). This suggests that overall quality may degrade due to the effects of a few heads dominating the outcome, indicating further room for improvement. To test this, we filtered and perturbed only high-scoring heads based on preference scores. The resulting samples (Fig.~\ref{fig:head-per-layer} (c)) show clear improvements over standard layer-level guidance. 
These results demonstrate the inefficiency of coarse layer-level guidance and motivate us to \textit{analyze head-level guidance and its properties}.

\begin{figure}[!t]
  \centering
  \includegraphics[width=\linewidth]{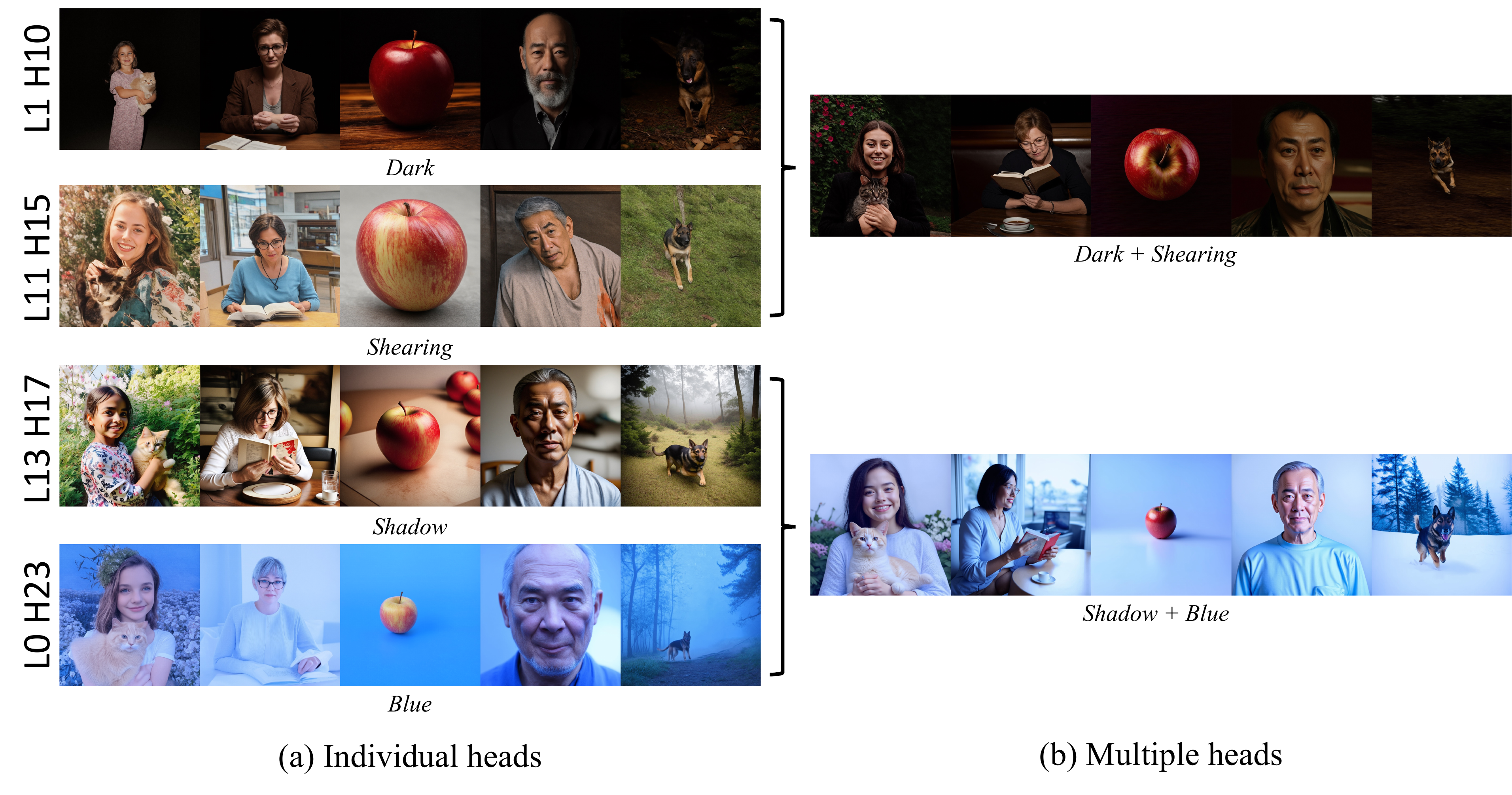}
  \caption{%
    \textbf{Effect of head-level guidance on concept amplification and combination.}
    (a) Guiding with individual heads amplifies specific visual concepts such as darkness, geometry, shadow, or color. 
    (b) Guiding with two heads simultaneously combines their effects in the output.}
  \vspace{-15pt}
  \label{fig:special-heads}
\end{figure}

\paragrapht{Individual head-level guidance occasionally reveals interpretable concepts.}
As can be seen in Fig.~\ref{fig:motivation}, we observe that perturbing individual heads for guidance leads to highly diverse behaviors in terms of their influence on generation. Beyond quality, we find that certain heads amplify specific visual attributes when used for guidance. These include texture, color tone, geometry, and lighting. We provide some examples in Fig.~\ref{fig:special-heads}. \texttt{(L11,H15)} consistently induces shearing, while other heads amplify darkness or blue tone. Additional interpretable heads and their effects are shown in Appendix~\ref{sup:subsec:interpretable_heads}. We further analyze these effects in depth in the discussion section, where we examine how individual heads contribute to image generation, how their roles differ across layers, and how they interact when composed together.







\paragrapht{Concepts can be composed via head combinations.}
Interestingly, combining multiple heads for head-level guidance results in the composition of their associated visual concepts. As shown in Fig.~\ref{fig:special-heads} (b), the generated images exhibit hybrid effects—such as the combination of lighting (\texttt{L1}, \texttt{H10}) and shearing distortions oriented from top-left to bottom-right (\texttt{L11}, \texttt{H15})—suggesting that head-level guidance can serve as a mechanism for concept-level control. More examples of concept composition are provided in Appendix~\ref{sup:subsec:more_combinations}.

\paragrapht{Composing heads increases image quality but may lead to over-saturation and over-simplification.}
Perturbing more heads strengthens the guidance effect but can also introduce undesirable artifacts such as oversaturation or oversimplification.  
As shown in Fig.~\ref{fig:head-numbers} (a), adding more perturbed heads results in over-perturbation, leading to over-smoothed or overly simplified output. Fig.~\ref{fig:head-numbers} (b) further illustrates this effect quantitatively: the quality metric (FID~\cite{kirstain2023pick}) using 1K prompts from the MS COCO dataset begins to degenerate as more heads are added after some threshold. This suggests that, in addition to the choice of perturbation method, the number of perturbed heads itself is a key factor for controlling the overall perturbation strength.

\section{Controlling Head-Level Attention Perturbation}
\label{sec:unified-perspective-attention-perturbation}

\vspace{-4pt}
\subsection{HeadHunter: An Iterative Framework for Retrieving Effective Attention Heads}
\label{subsec:headhunter}

The analysis in Sec.~\ref{subsec:head-properties} revealed that individual attention heads exert distinct and often complementary influences via head-level guidance—some modulate lighting, others control color or geometry. Moreover, these effects can be \textit{composed}, meaning that combinations of heads can yield richer and more controllable outputs than any single head alone. This raises a key question:
Can we guide image generation toward a \textit{user-defined} objective by selectively perturbing attention heads that are \textit{automatically} identified based on their individual and compositional effects?

To address this, we introduce \textbf{HeadHunter}, a framework that optimizes an arbitrary objective function by iteratively selecting a subset of attention heads to perturb. Each round consists of three stages: generation, evaluation, and expansion. In the generation stage, the framework perturbs attention maps for each candidate head $(l, h) \in \mathcal{S}$ and generates samples using attention perturbation guidance with multiple prompt–seed pairs $\mathcal{Q} = \{(p_1, s_1), \dots, (p_M, s_M)\}$. During the evaluation stage, it computes the average objective score using the user-given objective function $\mathcal{O}$ over the generated samples for each candidate. In the expansion stage, the top-$k$ performing heads are added to the selected head set $\mathcal{S}_{\text{final}}$. This process is repeated for a fixed number of rounds $R$. The full procedure is described in Algorithm~\ref{alg:headhunter-iter}. We validate HeadHunter both quantitatively and qualitatively across tasks such as general quality and style enhancement.

This iterative design greedily selects attention heads in the order that most increases the objective. As a result, it achieves rapid improvements during the early rounds (Fig.~\ref{fig:quanstyle-heads-a}). Later rounds enable the selection of heads that strengthen specific styles, such as imposing a particular tone on the image (Fig.~\ref{fig:suboptimal}), even if they do not improve overall quality. 
We present detailed experimental results and analysis in later sections.

\label{sup:subsec:headhunter-algorithm}
\begin{algorithm}
\caption{HeadHunter: Iterative Objective-Aware Head Selection}
\label{alg:headhunter-iter}
\begin{algorithmic}[1]
\Require Diffusion model $\mathcal{M}$, objective function $\mathcal{O}$, prompt-seed pairs $\mathcal{Q} = \{(p_1, s_1), \dots, (p_M, s_M)\}$, attention head set $\mathcal{S} = \{(l_1, h_1), \dots, (l_L, h_{H_L})\}$, number of heads per round $k$, number of rounds $R$
\Ensure Selected heads $\mathcal{S}_{\text{final}}$
\State $\mathcal{S}_{\text{final}} \gets \emptyset$ \Comment{Initialize selected head set}
\State $\mathcal{R} \gets \mathcal{S}$ \Comment{Initialize remaining head pool}
\For{$t = 1$ to $R$}
    \State $\mathcal{C} \gets [\,]$ \Comment{Initialize temporary score list}
    \For{each $(l, h) \in \mathcal{R}$}
        \State \textcolor{gray}{\# \textbf{Generation}}
        \State $\mathcal{S}_{\text{target}} \gets \mathcal{S}_{\text{final}} \cup \{(l, h)\}$ \Comment{Define perturbation set}
        \State Generate $\{\hat{x}_j\}_{j=1}^M$ using $\mathcal{M}$ with perturbation guidance on $\mathcal{S}_{\text{target}}$ and $(p_j, s_j) \in \mathcal{Q}$
        \State \textcolor{gray}{\# \textbf{Evaluation}}
        \State $s_{(l,h)} \gets \frac{1}{M} \sum_{j=1}^M \mathcal{O}(\hat{x}_j,p_j)$ \Comment{Compute average objective score}
        \State Append $((l, h), s_{(l,h)})$ to $\mathcal{C}$
    \EndFor
    \State \textcolor{gray}{\# \textbf{Expansion}}
    \State Sort $\mathcal{C}$ by $s_{(l,h)}$ in descending order
    \State $\mathcal{S}_{\text{new}} \gets$ top-$k$ heads from $\mathcal{C}$
    \State $\mathcal{S}_{\text{final}} \gets \mathcal{S}_{\text{final}} \cup \mathcal{S}_{\text{new}}$
    \State $\mathcal{R} \gets \mathcal{R} \setminus \mathcal{S}_{\text{new}}$
\EndFor
\State \Return $\mathcal{S}_{\text{final}}$
\end{algorithmic}
\end{algorithm}

\begin{figure}[!t]%
    \centering
\includegraphics[width=\linewidth]{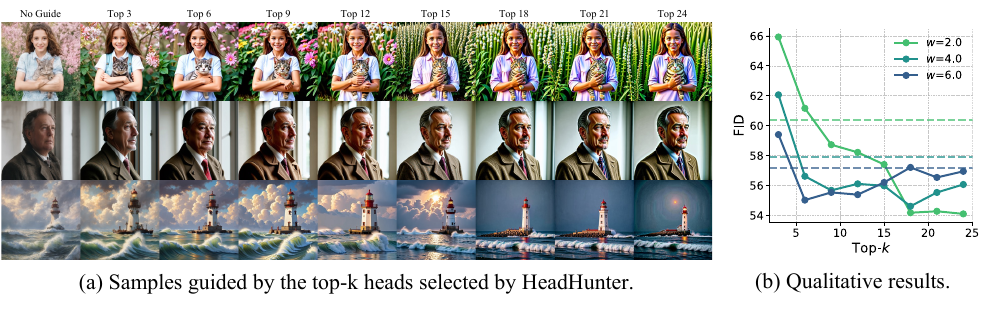}
    \vspace{-15pt}
    \caption{\textbf{Results of HeadHunter for general image quality improvement.} Performance improves as more top-ranked heads are added, demonstrating the effectiveness of compositional head selection via HeadHunter. The dotted horizontal line in (b) indicates the best score achieved by layer-level guidance, which is surpassed by a compact set of top-$k$ heads for $k < 10$. Dashed lines indicate the FID of the best-performing layer-level perturbation for each guidance scale $w$ in Eq.~\ref{eq:attention-perturbation-guidance}.}
    \label{fig:head-numbers}%
    \vspace{-5pt}
\end{figure}

\vspace{-4pt}
\subsubsection{Improving general image quality}
\label{subsec:headhunterquality}

In Sec.~\ref{subsec:head-properties}, we observe that even layers considered ineffective under layer-level guidance can improve quality when selectively activating appropriate heads. This suggests that \textit{fine-grained head-level selection across the entire set of heads} may yield stronger enhancements in image quality. To explore this, we apply HeadHunter to search the head selection space.

\paragraph{Experiments.}
Various metrics can be used to assess image quality, including vision-language models and learned reward functions. Following recent trends in diffusion models that optimize image preference scores~\cite{schuhmann2022laion, kirstain2023pick,xu2023imagereward,zhang2024learning}, we demonstrate our method using PickScore~\cite{kirstain2023pick} as the objective $\mathcal{O}$. HeadHunter is applied with 20 prompt-seed pairs $\mathcal{Q}$, one round ($R = 1$), and a selection of $k = 24$ heads per round. Additional implementation details are provided in Appendix~\ref{sup:sec:headhunter}. 
Qualitative results in Fig.~\ref{fig:head-numbers} (a) show that image quality progressively improves as more HeadHunter-selected heads are included in the guidance process, supporting the compositional nature of head-level perturbations observed earlier. Fig.~\ref{fig:head-numbers} (b) presents quantitative results in terms of FID, evaluated on 1K prompts from the MS COCO~\cite{lin2014microsoft} validation set. The performance improves as top-ranked heads are incrementally added, except in the high guidance scale case ($w = 6.0$), where oversaturation may occur.
Remarkably, using only 25\% of the heads ($k = 6$ out of 24) achieves comparable to or even better performance than full layer-level perturbation (shown as dotted lines for each guidance scale) and can be boosted much further by using additional heads. These results suggest that HeadHunter can identify a compact yet effective subset of attention heads, outperforming heuristic layer-based selection strategies. 


\begin{figure}[!t]
    \centering
    \includegraphics[width=\linewidth]{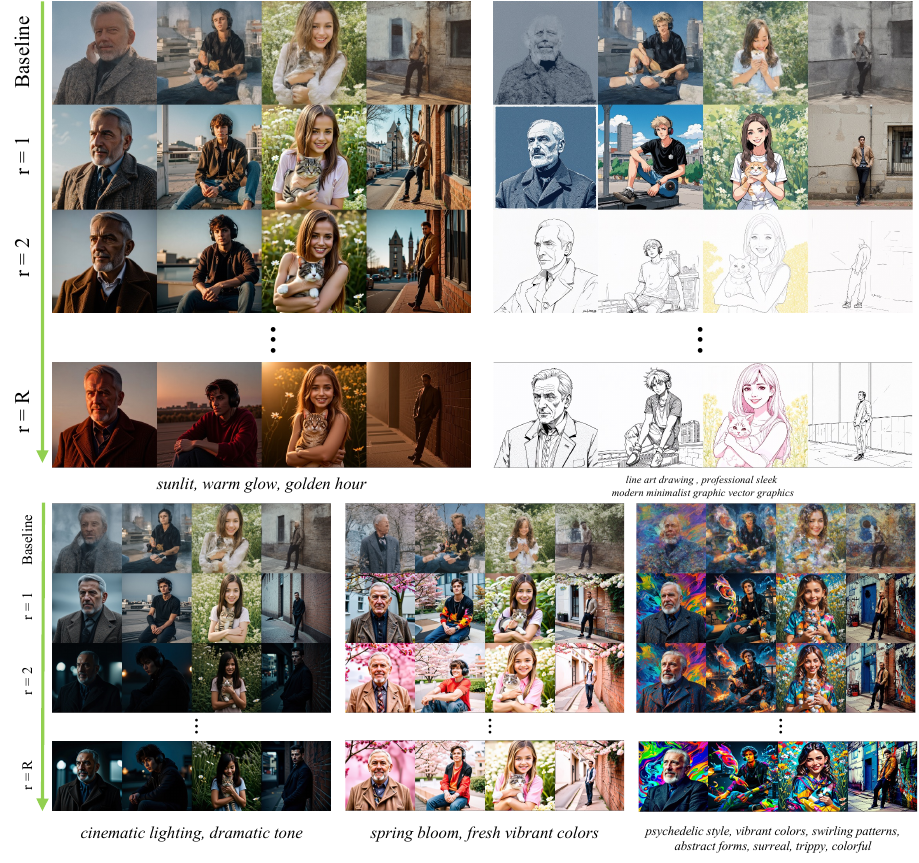}
    \caption{\textbf{Qualitative results of HeadHunter for style-oriented quality improvement.} The first row presents the unguided result. In each subsequent row, we apply head-level guidance using additional attention heads that are incrementally selected by the HeadHunter framework. As more heads are accumulated over rounds, the generated images exhibit progressively stronger alignment with the target style while improving visual quality. Original figures and additional results conducted in FLUX.1-Dev~\cite{blackforestlabs_flux1_dev} can be found in Appendix~\ref{sup:subsec:headhunter-impl-additional-restuls}.}
    \label{fig:qualstyle-heads}
    \vspace{-10pt}
\end{figure}

\vspace{-4pt}
\subsubsection{Improving style-oriented quality}
\label{subsec:headhunterstyle}

While Section~\ref{subsec:headhunterquality} demonstrated HeadHunter’s effectiveness in improving 
overall fidelity of samples, many real-world applications require more targeted control over \textit{specific} visual styles, such as evoking a particular mood, or mimicking classical art techniques. Inspired by the findings of Sec.~\ref{subsec:head-properties} and Fig.~\ref{fig:special-heads}, which reveal that certain attention heads are responsible for distinct stylistic or geometric attributes, we investigate whether HeadHunter can selectively enhance a target style through head-level guidance—while preserving or even enhancing overall image quality. We refer to this approach as style-oriented quality improvement.


\begin{wrapfigure}[15]{r}{0.4\linewidth}
    \vspace{-15pt}
    \centering
    \includegraphics[width=\linewidth]{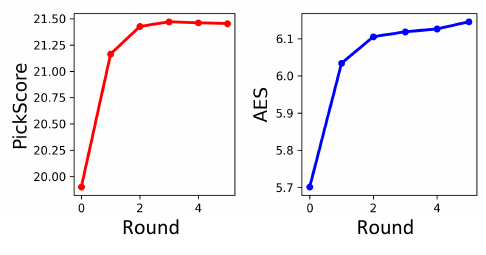}
    \vspace{-15pt}
    \caption{\textbf{Quantitative results of HeadHunter for style-oriented quality improvement.} As more heads accumulate, the generated images progressively align better with the target style and exhibit improved visual quality.}
    \label{fig:quanstyle-heads-a}
    \vspace{-5pt}
\end{wrapfigure}

\vspace{-5pt}
\paragraph{Experiments.} To evaluate HeadHunter’s ability to enhance specific styles, we prepare two prompt sets: one for \textit{style} (e.g., “warm golden hour glow”) and one for \textit{content} (e.g., “portrait of a violinist”). Each trial uses a composite prompt of the form “\texttt{style, content}.” We run HeadHunter with $R = 5$ and $k = 3$. Fig.~\ref{fig:qualstyle-heads} shows that, as more heads are selected, the generated images increasingly exhibit the intended style while maintaining structural integrity.

Fig.~\ref{fig:quanstyle-heads-a} reports quantitative metrics: PickScore~\cite{kirstain2023pick} and the LAION Aesthetic Score (AES)~\cite{schuhmann2022laion} both improve steadily as $k$ increases in the prompt-seed pairs $\mathcal{Q}$. These results confirm that HeadHunter effectively amplifies target styles in alignment with human preferences. For additional qualitative examples with SD3~\citep{esser2024scaling} and FLUX.1-Dev~\citep{blackforestlabs_flux1_dev}, we refer readers to \Cref{fig:qual_sd3_sunlit_spring,fig:qual_sd3_cinematic_psychedelic,fig:qual_sd3_line_flat,fig:qual_flux_acid_cubist,fig:qual_flux_cyberpunk_flat,fig:qual_flux_spring_stained} in Appendix.

\begin{figure}[!t]
    \centering
    \begin{minipage}[t]{\linewidth}
     \centering
        \captionof{table}{\textbf{Quantitative evaluation of generalizability to unseen content prompts.} Number in the parenthesis denotes guidance scale $w$. Applying HeadHunter (style-oriented quality setting) to unseen content prompts demonstrates strong generalization, yielding significantly higher human preference scores than the baseline and performance comparable to CFG.}
        \resizebox{0.8\textwidth}{!}{
        \begin{tabular}{l|cccc}
            \toprule
            \textbf{Method} & \textbf{PickScore} $\uparrow$ & \textbf{AES} $\uparrow$ & \textbf{HPS} $\uparrow$ & \textbf{Imreward} $\uparrow$\\
            \toprule
            Baseline & 19.66 & 5.37 & 0.2147 & -0.591 \\
            \midrule
            CFG (3.0) & 20.87 & 5.71 & 0.2924 & 0.844 \\
            CFG (6.0) & 20.92 & 5.80 & 0.3046 & 1.063 \\
            \midrule
            HeadHunter (3.0) & 20.70 & 5.92 & 0.2901 & 0.470 \\
            CFG (3.0) + HeadHunter (3.0) & 20.92 & 5.93 & 0.3036 & 0.845 \\
            \bottomrule
        \end{tabular}}
        \label{tab:cfg_comaprison_sd3}
        \vspace{5pt}
    \end{minipage}
    
    \begin{minipage}[t]{\linewidth}
        \centering
        \includegraphics[width=\linewidth]{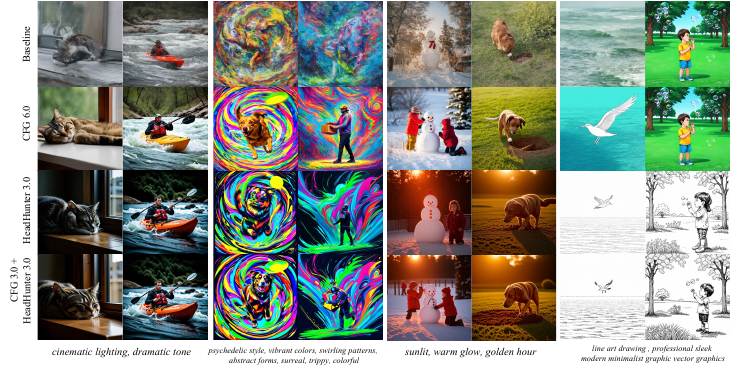}
        \caption{\textbf{Qualitative evaluation of generalizability to unseen content prompts.} Applying HeadHunter (style-oriented quality setting) results in substantial enhancement of stylistic attributes, outperforming not only the baseline but also CFG.}
        \label{fig:cfg_comparison_sd3}
    \end{minipage}
        
    \begin{minipage}[t]{\linewidth}
        \centering
        \includegraphics[width=\linewidth]{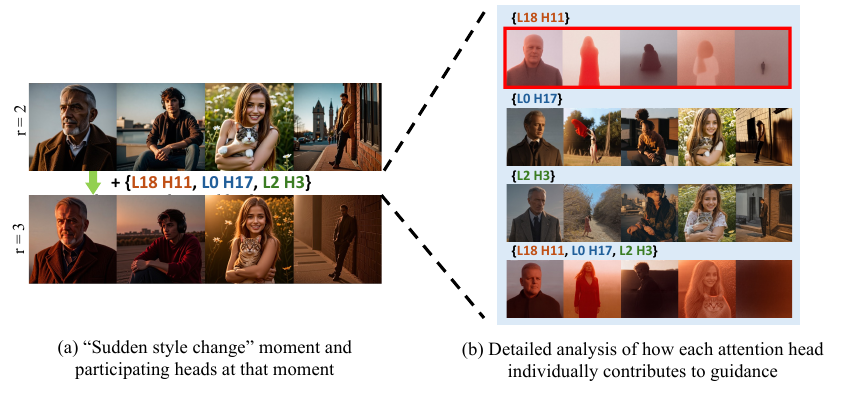}
        \caption{\textbf{Role of individually weak heads.} For sudden stylistic transition moments (e.g., adding global warmth in (a)), we visualize the effect of newly added heads in (b). One of these heads generate blurry reddish outputs when used alone (see red box). Although they fail to produce meaningful content when used independently, they contribute effectively when composed with previously selected structural heads. This highlights the importance of iterative strategy since those heads are unlikely to be selected by one-shot evaluation due to low quality. Additional results can be found in Fig.~\ref{fig:suboptimal_full}}
        \label{fig:suboptimal}
        
    \end{minipage}
    \vspace{-20pt}
\end{figure}

\paragraph{Assessing generalizability to unseen content prompts.}
To evaluate the generalizability of HeadHunter, we compare it against both the baseline and CFG using a set of 50 unseen content prompts. As shown in Tab.~\ref{tab:cfg_comaprison_sd3}, HeadHunter achieves significantly higher human preference scores than the baseline and performs comparably to CFG, thereby validating its ability to generalize to novel content prompts. Notably, as illustrated in Fig.~\ref{fig:cfg_comparison_sd3}, \textbf{HeadHunter leads to a substantial enhancement of stylistic attributes even compared to CFG}. For the style prompt \textit{sunlit, warm glow, and golden hour}, HeadHunter produces visibly intensified reddish tones and sunlight effects. Likewise, for the \textit{line art drawing ...} style prompt, monotone line characteristics are markedly enhanced. These results suggest that HeadHunter can serve as an effective plug-and-play module within existing inference pipelines (including those employing CFG) to improve both stylistic fidelity and overall image quality without requiring additional training.

\subsubsection{Discussion}

\paragraph{Surprising utility of individually weak heads.}
\label{sup:para:suboptimal-heads}
Interestingly, some heads that individually produce poor outputs can still play an essential role in composition. In Fig.~\ref{fig:suboptimal} (a), we observe a sudden enhancement in stylistic expression (e.g., intensified red) with inclusion of certain head set. By individually performing head-level guidance with each head in the set, we found that some head alone yields low-quality outputs but it reliably enforces a style trait. Such heads are unlikely to be selected through one-shot evaluation due to low quality, but HeadHunter's iterative nature allows them to be integrated later as their utility emerges in composition. This highlights HeadHunter’s capacity to exploit \textit{compositional synergies} among heads.

\subsection{Controlling Guidance Intensity via Attention Map Interpolation}\label{subsec:softpag}

Although our head-level guidance provides fine-grained, objective-specific control, we propose an orthogonal approach that enables \textit{continuous} modulation of the guidance effect. This allows for more precise adjustment of the perturbation intensity applied to selected heads or layers. Specifically, we introduce a simple interpolation strategy that interpolates the original attention map $\mathbf{A}$ with its perturbed counterpart. For example, the interpolated attention map for perturbed-attention guidance (PAG)~\cite{ahn2024self} is defined as:
\begin{equation}
\label{eq:identity-interpolation}
\mathbf{A}_{l, h}^{(\text{SoftPAG})} = (1-u) \mathbf{A}_i + u \mathbf{I}, \quad \text{ for } (l, h) \in \mathcal{S}, \quad u \in [0, 1].
\end{equation}
This formulation replaces $\mathbf{A}_{l,h}^{\text{(PAG)}}$ in Eq.~\ref{eq:pag} with its interpolated version, enabling a smooth transition between the original attention map and the fully perturbed one.

\begin{figure}[t]
    \centering
    \includegraphics[width=\linewidth]{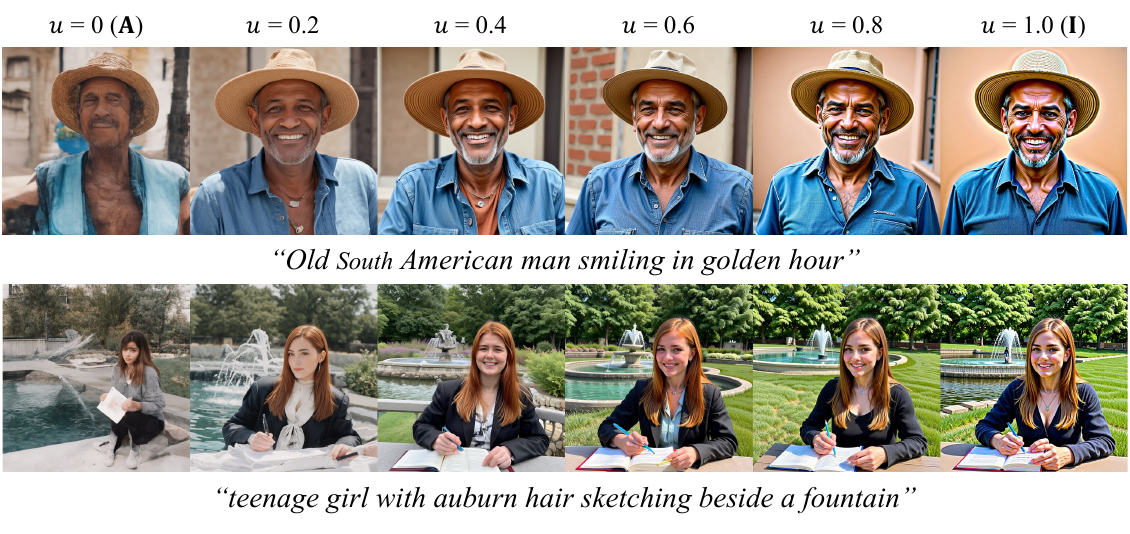}
    \vspace{-10pt}
    \caption{\textbf{Generated Images with linear interpolation between attention map $\attnmap$ and an identity matrix $\mathbf{I}.$} While increasing $u$ enhances quality up to a point, it eventually results in over-saturation and structural over-simplification. Detailed hyperparameters can be found in Appendix~\ref{sup:sec:experimental-details}.}
    \label{fig:AI_interpolation}
    \vspace{-15pt}
\end{figure}

We present qualitative results across varying $u$ values in Fig.~\ref{fig:AI_interpolation}. These results show that increasing the interpolation parameter $u$ initially enhances sample quality by rectifying structural artifacts in the samples. However, beyond a certain point, it leads to oversimplified images with smoothed backgrounds and exaggerated structures. The interpolation method provides a \textbf{``sweet spot''} where the samples exhibit improved structure while retaining visually realistic features. By treating the identity-matrix perturbation in PAG~\cite{ahn2024self} as a specific point in probability distribution space, our formulation enables a smooth and principled interpolation between $\mathbf{A}$ and $\mathbf{I}$. The scalar $u \in [0,1]$ provides an intuitive and interpretable control over perturbation strength. Additionally, since this method is a simple linear interpolation of self-attention maps
, it can be implemented with a few lines of code. We refer to this controllable variant of PAG as \textbf{Soft Perturbed-Attention Guidance (SoftPAG)}. 

Note that this interpolation approach represents a general principle that can be applied to modulate the strength of various attention-based guidance methods, including existing ones such as Smoothed Energy Guidance~\cite{hong2024smoothed} and interpolations toward a constant matrix (i.e., uniform attention weights). A broader discussion of the unified perspective on attention perturbation and corresponding results is provided in Appendix~\ref{sup:sec:interpolate-quan}.

\paragraph{Interpolation parameter $u$ \textit{vs.} guidance scale $w$.}

Both the guidance scale $w$ and interpolation parameter $u$ influence the strength of perturbation, often affecting saturation and structural fidelity. This raises the question of whether adjusting $u$ offers any advantage over simply tuning $w$.

Fig.~\ref{fig:interpolation-parameter-vs-guidance-scale-pag} in Appendix visualizes metric values over a grid of $(w, u)$ pairs, where brighter regions indicate better performance and red boxes denote optimal configurations. \textbf{Across most metrics, the best-performing setting occurs at $u < 1.0$,} indicating that full replacement (i.e., $u=1$ as in PAG) is often suboptimal. These results underscore the importance of explicitly controlling perturbation strength via interpolation, rather than relying solely on the guidance scale. Due to space limitations, we defer a detailed discussion of the effectiveness of SoftPAG and the selection of $u$ and $w$ to Appendix~\ref{sup:sec:softpag_hyperparam}.

\section{Conclusion}

In this work, we move beyond heuristic layer selection in attention perturbation guidance by identifying attention heads as a more meaningful and fine-grained unit of intervention. To the best of our knowledge, this is the first work to perform attention perturbation at the level of individual heads. Specifically, we present HeadHunter, an iterative framework for selecting semantically relevant attention heads aligned with arbitrary, user-defined objectives. Empirical results on state-of-the-art DiT-based models, including Stable Diffusion 3 and FLUX.1, show that head-level perturbation not only improves general image quality with a single search but also aesthetically enhances specific visual styles. In addition, we introduce SoftPAG, a lightweight yet powerful mechanism for continuously modulating perturbation strength via attention map interpolation. Together, these two approaches address key limitations of prior guidance methods, enabling more targeted, effective, and controllable inference-time interventions. We believe this work opens promising directions for interpretable and modular control in generative modeling.


\begin{ack}

This research was supported by Institute of Information \& communications Technology Planning \& Evaluation (IITP) grant funded by the Korea government (MSIT) (RS-2019-II190075, RS-2024-00509279, RS-2025-II212068, RS-2023-00227592, RS-202502214479, RS-2024-00457882, RS-2025-25441838, RS-2025-25441838, RS-2025-02214479, RS-2025-02217259) and the Culture, Sports, and Tourism R\&D Program through the Korea Creative Content Agency grant funded by the Ministry of Culture, Sports and Tourism (RS-2024-00345025, RS-2024-00333068, RS-2023-00222280, RS-2023-00266509), and National Research Foundation of Korea (RS-2024-00346597).
\end{ack}


\bibliographystyle{plain}
\bibliography{neurips_2025}     

\clearpage
\appendix

\setcounter{page}{1}

\section*{\Large Appendix}



This appendix provides supplementary material to support the main paper.

\vspace{-50pt}
\mtcsettitle{parttoc}{}
\part{}
\section*{Contents}
\parttoc

\section{Related Works}
\label{sup:related_works}

\paragraph{Diffusion models.}
Diffusion models~\cite{ho2020ddpm, Song2020ddim, jascha2015, yang2019estgrad, yang2020sde, dhariwal2021diffusion, rombach2022high, dustin2023sdxl} have become the foundation of modern generative modeling, driving advances in both image~\cite{rombach2022high, dustin2023sdxl, peebles2023scalable, chen2024pixart, esser2024scaling} and video synthesis~\cite{ho2022video, Ho2022ImagenVH, Blattmann2023StableVD, Zhuoyi2025cogvideox}.
Early methods rely on stochastic differential equations (SDEs) to learn a reverse denoising process from noise to data.
More recently, deterministic samplers based on ordinary differential equations (ODEs), including rectified flow~\cite{nanye2024sit, esser2024scaling} and flow matching~\cite{Lipman2022FlowMatching}, have emerged as efficient alternatives, learning continuous trajectories that transform noise into data.
These methods accelerate convergence and improve stability, especially in large-scale models.
In parallel, network architectures have shifted from U-Net backbones~\cite{dustin2023sdxl} to Diffusion Transformers (DiT)~\cite{peebles2023scalable, chen2024pixart, esser2024scaling, chen2023pixart, chen2024pixartdelta}, enhancing scalability and representational capacity.

\paragraph{Attention perturbation guidance.}
In recent studies, various modifications and extensions of classifier-free guidance have been introduced. Among them, methods that directly perturb the attention layers for guidance work effectively. Self-Attention Guidance (SAG)~\cite{hong2023improving} applies Gaussian blur to the self-attention layers of the UNet to introduce perturbations. Similarly, Perturbed-Attention Guidance (PAG)~\cite{ahn2024self} generates perturbations by replacing the attention maps with an identity matrix. Spatiotemporal Skip Guidance (STG)~\cite{hyung2025spatiotemporal} is a 3D extension of PAG that improves sample quality by selectively skipping spatiotemporal layers. Smoothed Energy Guidance (SEG)~\cite{hong2024smoothed} defines the energy of self-attention and employs 2D Gaussian blur to reduce the curvature of the energy landscape, using the result as a form of perturbation.  Autoguidance~\cite{tero2024} enhances image quality by using a bad version of the same conditional model as a source of perturbation. Self-Guidance (SG)~\cite{li2024self} leverages the model prediction at a noisy timestep to generate perturbations.




\clearpage
\section{Continuous control on perturbation strength via attention map interpolation}
\label{sup:sec:entropy-perturbation}

\label{sup:sec:softpag_hyperparam}
\begin{figure}[!h]
    \centering
    \includegraphics[width=\linewidth]{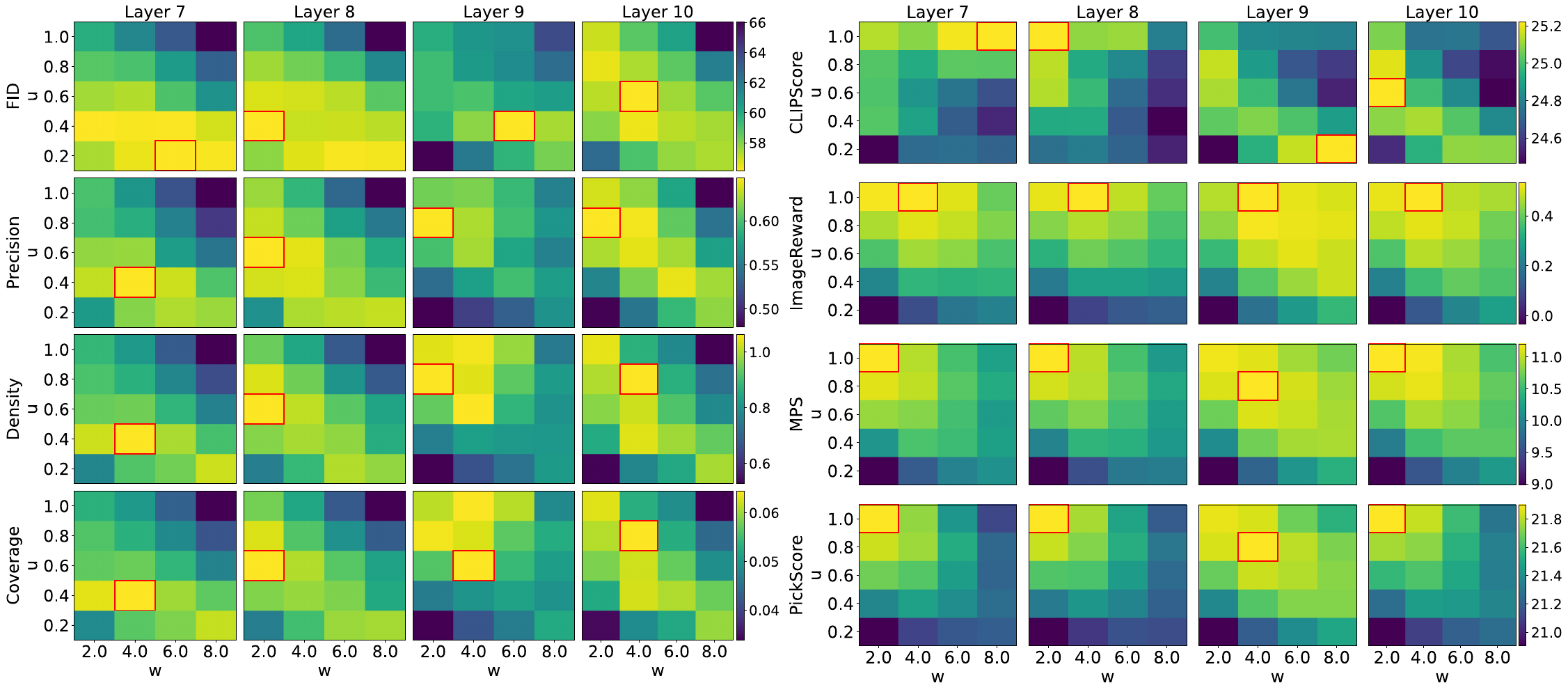}
    \caption{\textbf{Grid search for SoftPAG, on guidance scale $w$ and interpolation parameter $u$ with different metrics.}}
    \label{fig:interpolation-parameter-vs-guidance-scale-pag}
\end{figure}

\begin{figure}[!h]
    \centering
    \includegraphics[width=0.5\linewidth]{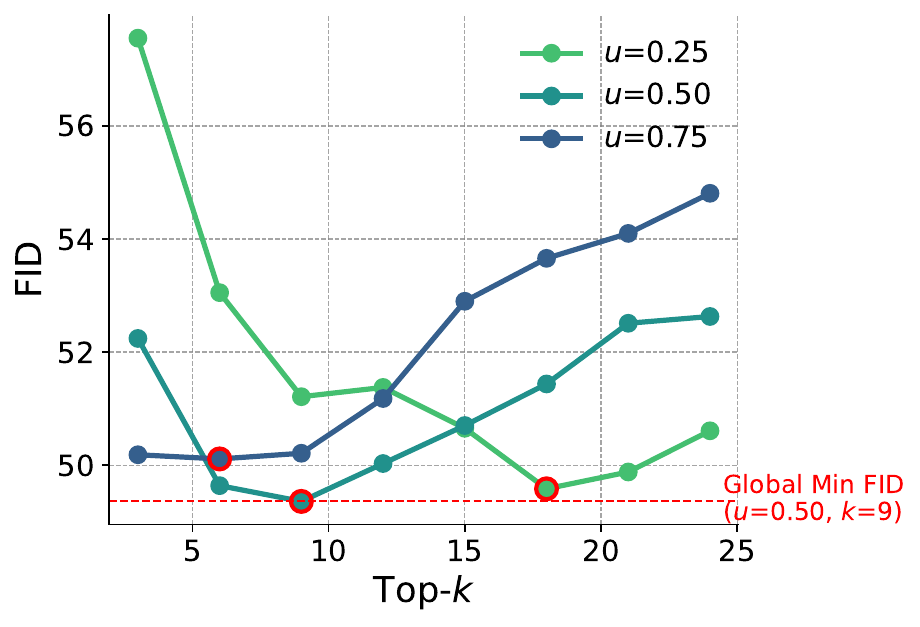}
    \caption{\textbf{Effect of interpolation parameter $u$ on FID across PickScore top-$k$ head selections.} 
    As $u$ decreases, the perturbation becomes milder, shifting the optimal point (lowest FID) toward larger $k$ (i.e., more heads are needed). 
    Notably, $u = 0.5$ yields the best FID (indicated by the red dashed line), highlighting the benefit of moderate perturbation strength.}
    \label{fig:u-vs-k}
\end{figure}

\paragraph{Pre-selecting interpolation parameter $u$ in HeadHunter search.}



As running HeadHunter multiple times with different interpolation parameters $u$ is computationally expensive, it is generally sufficient to search for heads using the original PAG. Nonetheless, we can explore the effect of $u$ through a single-round search to study general quality trends. To this end, we run HeadHunter with the same setup as in Sec.~\ref{subsec:headhunterquality}, replacing PAG with SoftPAG (Eq.~\ref{eq:identity-interpolation}) and varying $u$. The results, shown in Fig.~\ref{fig:u-vs-k}, report FID~\cite{martin2017} scores as a function of $k$ for different $u$ values.

We observe that lower $u$ values (i.e., softer perturbations) require more heads to achieve optimal performance, but also tend to yield better overall results with a moderate setting. Empirically, we find that $u = 0.5$ and $k = 9$ offer a strong trade-off.

This supports the view that both $u$ and $k$ modulate the strength of guidance. While this paper introduces two axes of fine-grained controllability—head-level selection and attention map interpolation—tuning $u$ is significantly more efficient than rerunning head selection. In practice, post-selecting $u$ after a standard head search is typically sufficient. However, if more compute is available, one may rerun HeadHunter with a moderate $u$ (e.g., $u = 0.5$) for a refined selection.

\begin{figure}[!h]
    \centering
    \includegraphics[width=\linewidth]{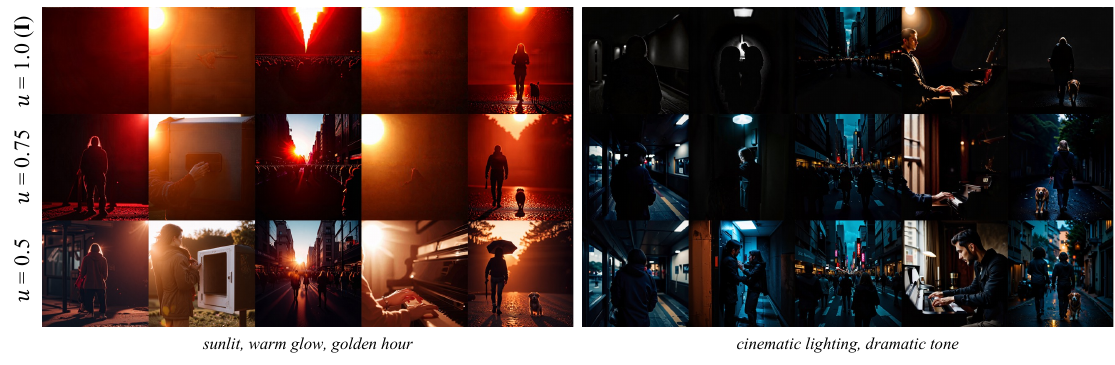}
    \caption{\textbf{Incorporating SoftPAG with the final heads retrieved by HeadHunter for style-oriented quality improvement.} Setting the interpolation parameter $u = 1.0$ makes the effect of head-level guidance more visible, which helps in selecting effective heads via HeadHunter. However, it may also introduce unwanted artifacts such as oversaturation or oversimplification. Post-tuning $u$ after head selection effectively reduces these artifacts while preserving the style enhancement, with minimal additional cost.}
    \label{fig:headhunter_softpag}
    \vspace{-5pt}
\end{figure}

\paragraph{Post-selecting of $u$ for artifact removal with retrieved heads.}
As discussed in Sec.~\ref{subsec:head-properties}, the number of heads used in head-level guidance affects how strong the perturbation is. When more heads are used, the intended effect such as style enhancement can be amplified. However, this may also lead to unwanted side effects including oversaturation or oversimplification due to bias of reward models~\cite{kirstain2023pick,schuhmann2022laion}. In the HeadHunter framework, heads are added step by step. This helps reinforce the target objective, but may also increase the risk of artifacts. To better control the strength of guidance, we can use SoftPAG, which allows us to mitigate the artifacts through the interpolation parameter $u$.

As tuning both the number of heads and the interpolation parameter $u$ at the same time can be complex and computationally expensive, we adopt a simple two-stage approach. First, we use HeadHunter with $u = 1.0$ to select effective heads. A larger $u$ helps make each head’s effect more visible, which is helpful during selection. Then, in the second stage, we optionally reduce $u$ to fine-tune the strength of the guidance and reduce any unwanted artifacts.

This approach offers a practical trade-off by eliminating complex tuning while retaining control over the final output. In Fig.~\ref{fig:headhunter_softpag}, we show examples where retrieved heads with $u = 1.0$ introduce unnatural artifacts. Reducing the SoftPAG parameter to $u = 0.5$ effectively removes these artifacts while preserving the style enhancement provided by the selected heads.

\clearpage
\section{Other perturbation methods in our framework and comparison}
\label{sup:sec:interpolate-quan}

\begin{figure}[!h]
    \centering
    \includegraphics[width=\linewidth]{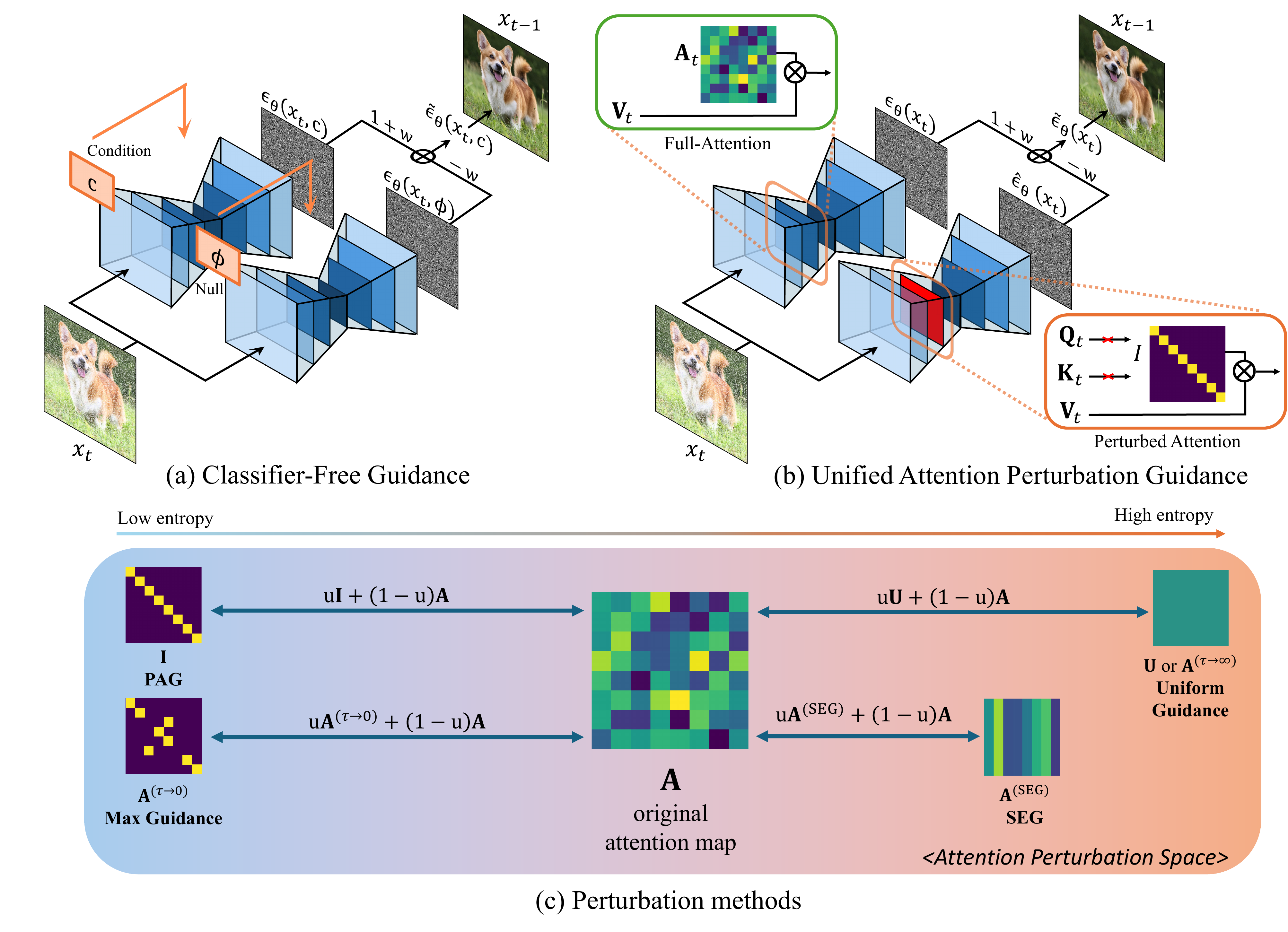}
    \caption{\textbf{Unified attention perturbation guidance.} Our framework unifies a variety of perturbation strategies by interpreting attention maps as probability distributions and interpolating between the original and perturbed variants.}
    \label{fig:unified-attention-perturbation}
\end{figure}

In the main paper, we focus on a single representative perturbation method: identity matrix replacement (PAG). Within our framework, which interprets the attention map as a probability distribution, identity perturbation can be seen as one end of the entropy-modulating transform that minimally shifts the distribution's entropy. Fig.~\ref{fig:unified-attention-perturbation} provides a conceptual overview of this unified view, with other perturbations. Our key insight is twofold: \textbf{(1) attention perturbations can be understood as transformations over probability distributions}, and \textbf{(2) the strength of any perturbation can be smoothly controlled via interpolation with the original attention map}. This enables elegant integration of existing methods while offering controllability.

\subsection{Uniform Guidance}

\begin{figure}[!h]
    \centering
    \includegraphics[width=\linewidth]{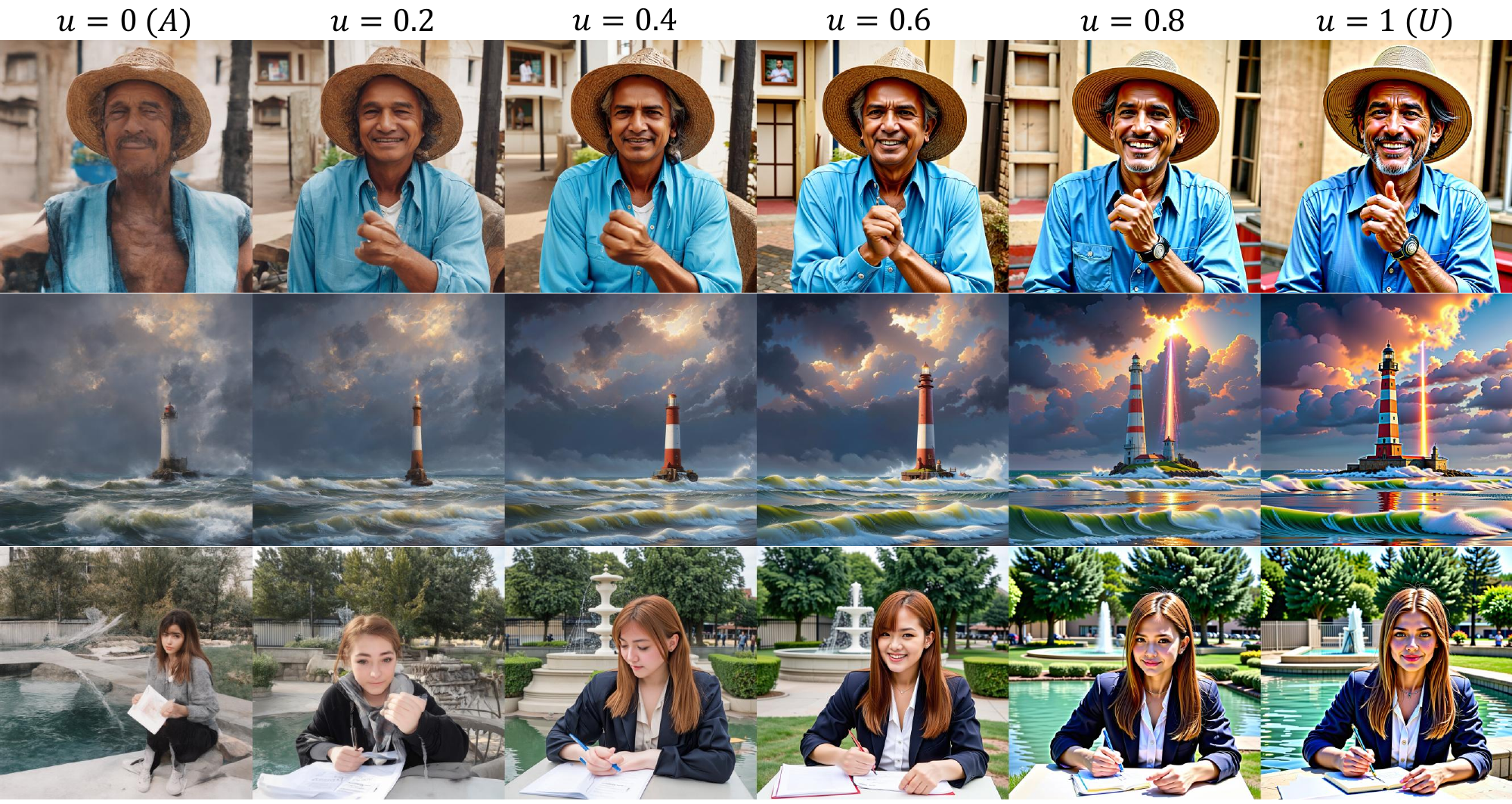}
    \vspace{-10pt}
    \caption{\textbf{Generated Images with linear interpolation between attention map $\attnmap$ and a uniform matrix $\mathbf{U}.$} While increasing $u$ enhances quality up to a point, it eventually results in over-saturation and structural over-simplification. Detailed hyperparameters can be found in Appendix~\ref{sup:sec:experimental-details}.}
    \label{fig:AU_interpolation}
    \vspace{-10pt}
\end{figure}

From an information-theoretic perspective, the identity matrix attention map is a special case where each row of the attention map corresponds to a Dirac delta distribution, which is the lowest possible entropy case. At the opposite extreme of the entropy spectrum lies the uniform distribution $\mathbf{U}$, whose rows are all \([\tfrac1N,\dots,\tfrac1N]\); it attains the
\emph{maximum} entropy, allowing each query to attend equally to every key. Replacing $\mathbf{A}$ by $\mathbf{U}$ in selected heads gives \textbf{Uniform Guidance (UG)}. Similar to the interpolation with the identity matrix in SoftPAG, we can define a distributional perturbation between the original attention matrix $\attnmap$ and $\mathbf{U}$ via linear interpolation:

\begin{equation}
\label{eq:uniform-interpolation}
\mathbf{A}^{(\text{SoftUG})} = (1-u) \mathbf{A} + u \mathbf{U}, \quad u \in [0, 1].
\end{equation}

We show the interpolation results in Fig.~\ref{fig:AU_interpolation}. Interestingly, this perturbation strategy tends to preserve the original structure of objects in the unguided sample ($u = 0$) slightly better. However, it often over-restores artifacts present in the original structure, sometimes generating unintended elements (e.g., a framed portrait in the background of Row 1, or a swimming pool in Row 3). Both methods, however, may exhibit oversaturation or oversimplification when overly strong perturbations are applied, highlighting the importance of a balanced interpolation with the original attention map $\mathbf{A}$.

\subsection{Smoothed Energy Guidance}
Smoothed Energy Guidance (SEG)~\cite{hong2024smoothed} applies a 2D Gaussian blur along the query axis to smooth the attention logits. By adjusting the kernel width $\sigma$, users can naturally control the strength of the perturbation. In the extreme case where $\sigma \to \infty$, SEG reduces to averaging all query features and applying the result globally.

\paragraph{SEG is not directly applicable to MMDiT.}
However, SEG is not directly compatible with MMDiT~\cite{esser2024scaling}, where both image and text tokens participate in attention. Applying a 2D Gaussian blur over the query axis becomes nontrivial when attention spans across modalities, as in recent text-to-video models where multiple frames attend jointly. This issue becomes more pronounced with the rise of multimodal attention (MM-attention).

\paragraph{Restoring controllability to SEG using our framework.}

\begin{figure}
    \centering
    \includegraphics[width=\linewidth]{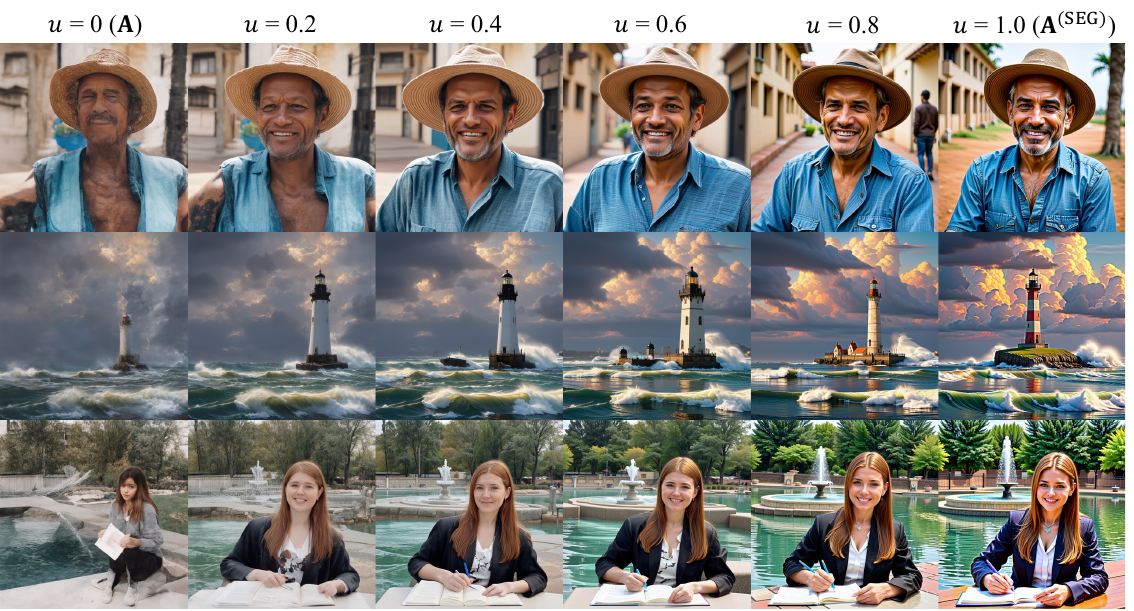}
    \caption{\textbf{Generated images with linear interpolation between attention map $\mathbf{A}$ and smoothed attention map $\mathbf{A}^{\text{(SEG)}}$}.}
    \label{fig:seg-interpolation}
\end{figure}
To address this, we propose a simple yet widely applicable solution: interpolate between the original attention map $\mathbf{A}$ and the one computed from the mean query feature—corresponding to the limiting case of SEG with $\sigma \to \infty$.




Specifically, we interpolate between the original attention map $\mathbf{A}$ and the SEG attention map $\mathbf{A}^{(\text{SEG})}$, computed using a mean query vector, as follows:
\begin{equation}
\mathbf{A^{(\text{SoftSEG})}} = (1 - u)\mathbf{A} + u\mathbf{A}^{(\text{SEG})}, \quad u \in [0, 1],
\end{equation}
where
\begin{equation}
\mathbf{A}^{(\text{SEG})} = \text{Softmax}\left( \frac{ \bar{\mathbf{Q}} \mathbf{K}^\top }{ \sqrt{d} } \right), \quad
\bar{\mathbf{Q}} := \mathbf{1}_N \bar{\mathbf{q}}^\top \in \mathbb{R}^{N \times d}, \quad
\bar{\mathbf{q}} := \frac{1}{N} \sum_{i=1}^{N} \mathbf{Q}_i \in \mathbb{R}^d.
\end{equation}
Here, $\mathbf{Q} \in \mathbb{R}^{N \times d}$ is the query matrix, and $\bar{\mathbf{Q}} \in \mathbb{R}^{N \times d}$ replicates the mean query vector across all $N$ tokens. This allows for a smooth transition from the token-specific attention map to a globally-averaged variant, effectively reintroducing controllability.

This provides a controllable and modality-agnostic extension of SEG to architectures like MMDiT, where traditional Gaussian blurring becomes infeasible. We show the interpolation results in Fig.~\ref{fig:seg-interpolation}.


\subsection{Softmax temperature scaling}
\paragraph{Attention perturbation using softmax temperature scaling.}
One intuitive way to increase or decrease entropy is by introducing a temperature parameter $\tau$ into the Softmax operation of attention, as follows:



\begin{figure}[h]
    \centering
    \includegraphics[width=\linewidth]{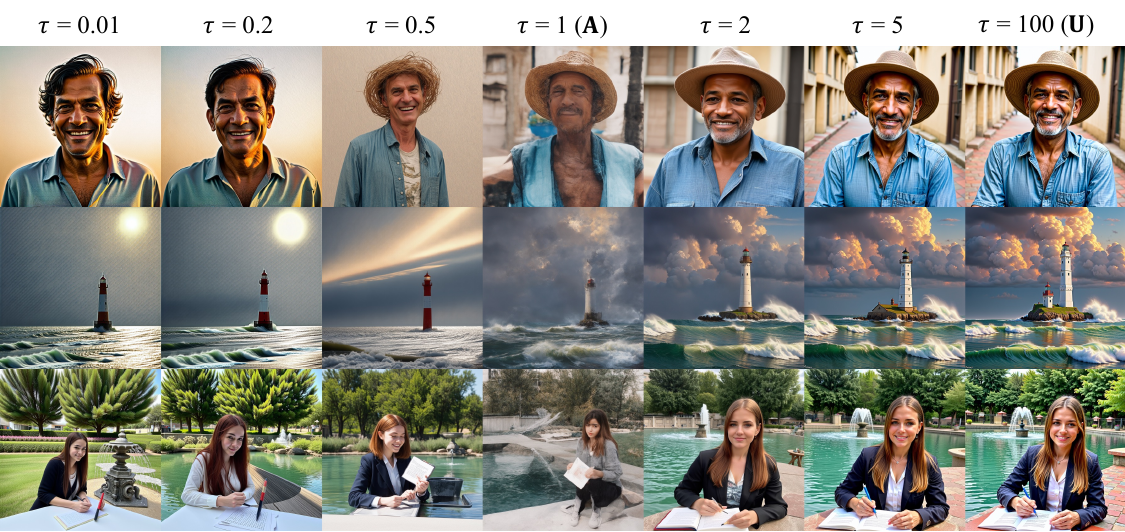}
    \caption{\textbf{Generated images with temperature scaling perturbation.}}
    \label{fig:T_interpolation}
\end{figure}

\begin{equation}
\mathbf{A}^{(\text{temp})} = \text{softmax} \left( \frac{\log \mathbf{A}}{\tau} \right), \quad \tau > 0.
\end{equation}

Lower temperatures ($\tau \to 0$) sharpen the distributions, driving them toward the Dirac delta distributions, while higher temperatures ($\tau \to \infty$) flatten the distribution towards uniform. This formulation provides a unified mechanism to traverse both entropy-increasing and entropy-decreasing directions within the attention perturbation space.


\paragraph{Perturbation $\tau \to 0$ results in different perturbation with PAG ($\mathbf{I}$).}
As $\tau \to \infty$, the distributions approach uniform distributions, following an exponential-geodesic path. In contrast, as $\tau \to 0$, the distribution converges to a Dirac delta distribution. However, unlike the identity matrix $\mathbf{I}$, where all probability mass is placed at each query's own index, this low-temperature limit concentrates mass on the maximum entry of each row, potentially differing from the diagonal. We illustrate the effect of varying $\tau$ in Fig.~\ref{fig:T_interpolation}. Notably, increasing and decreasing $\tau$ leads to meaningful quality improvements, suggesting that modulation of entropy in either direction can enhance generation.

\paragraph{Intuitive control parameter using interpolation framework.}
Softmax temperature scaling provides a simple way to modulate perturbation strength by adjusting the temperature $\tau \in (0, \infty)$. However, it introduces a practically unintuitive control parameter $\tau$, and the limiting distribution as $\tau \to 0$ is theoretically unreachable in practice due to numerical instability.

Our interpolation framework offers a compelling alternative: it allows for control using a normalized and intuitive parameter $u \in [0, 1]$, while enabling us to directly realize the theoretical limit of the attention map as $\tau \to 0$, denoted as $\mathbf{A}^{\tau \to 0}$, without the numerical issues associated with temperature scaling.

\begin{figure}
    \centering
    \includegraphics[width=\linewidth]{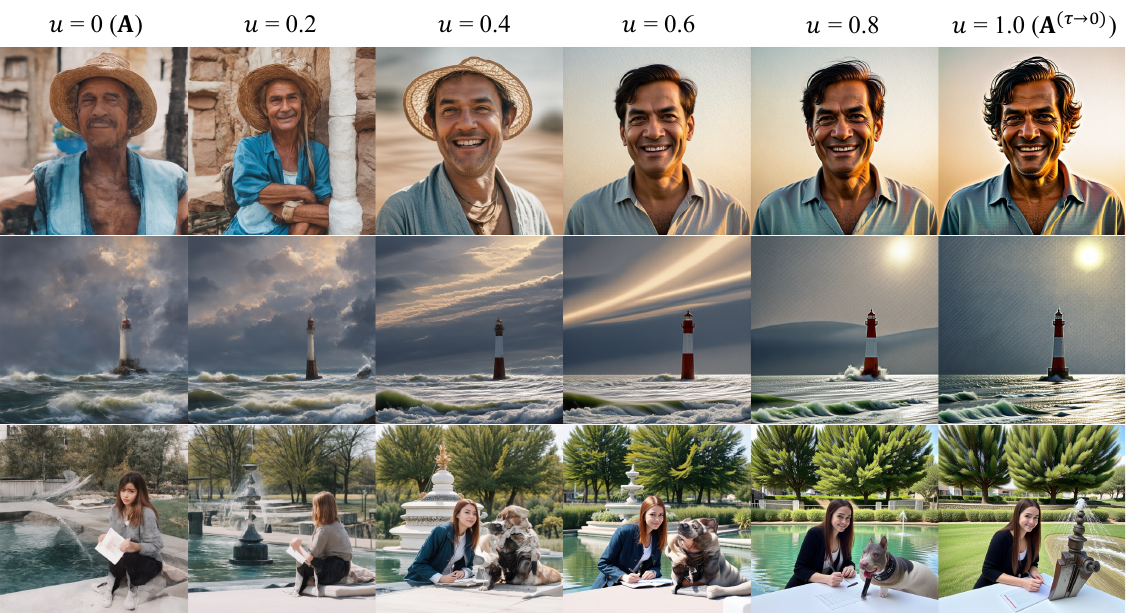}
    \caption{\textbf{Generated images with linear interpolation between attention map $\mathbf{A}$ and smoothed attention map $\mathbf{A}^{(\tau \to 0)}$}.}
    \label{fig:temperature-interpolation}
\end{figure}

Concretely, in the limit $\tau \to 0$, the softmax attention converges to a one-hot distribution where, for each row, the position of the maximum logit becomes 1 and all others become 0:
\begin{equation}
\mathbf{A}^{(\tau \to 0)}_{ij} :=
\begin{cases}
1 & \text{if } j = \arg\max_k \mathbf{A}_{ik} \\
0 & \text{otherwise}.
\end{cases}
\end{equation}
Then the perturbation methods is as follows:
\begin{equation}
\mathbf{A}^{(\text{SoftMG})} = (1 - u)\mathbf{A} + u\mathbf{A}^{({\tau \to 0})}, \quad u \in [0, 1].
\end{equation}
As shown in Fig.~\ref{fig:temperature-interpolation}, our framework can replicate this limiting behavior while preserving controllability through a continuous interpolation parameter. Compared to softmax temperature scaling (Fig.~\ref{fig:T_interpolation}), this approach offers a more intuitive and stable mechanism for navigating the space of attention perturbations. We refer to this perturbation-based guidance as \textbf{Max Guidance (MG)}.

\paragraph{Quantitative comparisons.}


We discuss various perturbation strategies and present their qualitative results in Fig.~\ref{fig:AI_interpolation},~\ref{fig:AU_interpolation},~\ref{fig:T_interpolation},~\ref{fig:seg-interpolation},~\ref{fig:temperature-interpolation}. To complement these findings, we provide quantitative comparisons in Tab.~\ref{tab:diversity-comparison}. 

Note that each perturbation method may benefit from different optimal hyperparameter configurations. To ensure a fair evaluation, we define a parameter pool and report the best-performing setting for each method and each metric. Specifically, we consider perturbation layers in the early-to-mid range, namely \texttt{layer 7}, \texttt{layer 8}, \texttt{layer 9}, and \texttt{layer 10}; guidance scales \(w \in \{2.0, 4.0, 6.0, 8.0\}\); and interpolation parameters \(u \in \{0.2, 0.4, 0.6, 0.8, 1.0\}\).

As shown in the Tab.~\ref{tab:diversity-comparison}, the $\textbf{I}$ perturbation consistently yields strong performance across most settings. Based on this robustness, we adopt it as our primary analysis tool in the main paper (Sec.~\ref{sec:analysis}, ~\ref{sec:unified-perspective-attention-perturbation}).

\begin{table}[H]
    \centering
    \small
    \resizebox{\linewidth}{!}{ 
    \begin{tabular}{l | c c c c c c c c}
        \toprule
        \textbf{Method} & \textbf{FID} $\downarrow$ & \textbf{Precision} $\uparrow$ & \textbf{Recall} $\uparrow$ & \textbf{Density} $\uparrow$ & \textbf{Coverage}  $\uparrow$& \textbf{PickScore} $\uparrow$& \textbf{ImageReward} $\uparrow$& \textbf{MPS} $\uparrow$\\
        \midrule
        Uniform Matrix $\mathbf{U}$  & 53.37 & 0.62 & \textbf{0.55} & 0.89 & 0.05 &21.90& 0.53&11.21\\
        Query Mean $\mathbf{A}^{(\text{SEG})}$  & \textbf{53.33} & 0.62 & 0.53 & 0.91 & \textbf{0.06}&21.89& 0.50& 11.13 \\
        Identity Matrix $\mathbf{I}$ & 56.15 & \textbf{0.65} & 0.48 & \textbf{1.06} & \textbf{0.06}&\textbf{22.19} &\textbf{0.65}&\textbf{11.70}\\
        \bottomrule
    \end{tabular}
    }
    \vspace{5pt}
    \caption{\textbf{Comparison of perturbation strategies across different metrics.} 
    Each method shows strengths in different metrics. Overall, Identity matrix perturbation (SoftPAG) achieves strong results across most quality, diversityand human preference metrics. Notably, identity matrix perturbation outperforms others in preference-based scores such as PickScore, indicating better alignment with human perception and more visually pleasing results. 
    Note that FID does not always correlate with human judgment.}
        \label{tab:diversity-comparison}
\end{table}

\paragraph{More quantitative results.}

\begin{figure}[!t]
    \centering
    \includegraphics[width=\linewidth]{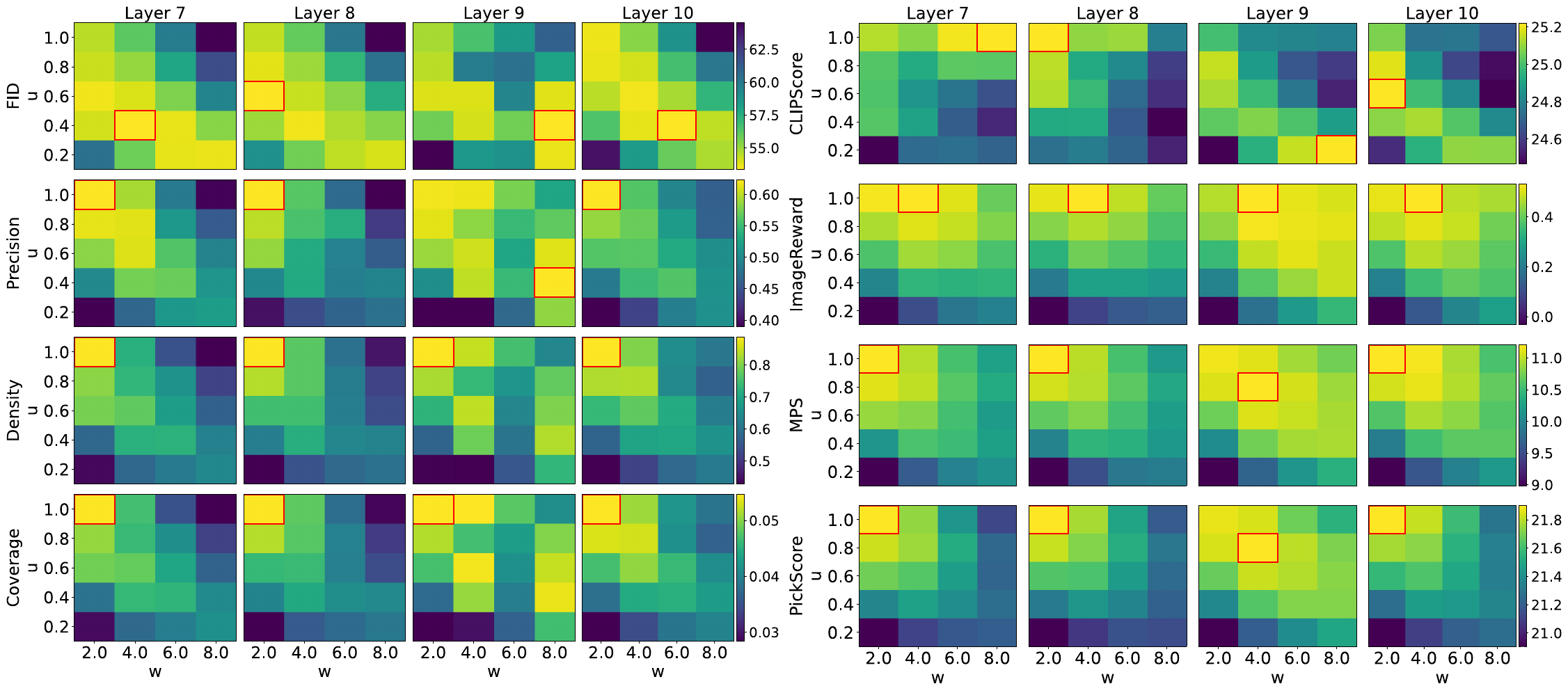}
    \caption{\textbf{Grid search for UG, on guidance scale $w$ and interpolation parameter $u$ with different metrics.}}
    \label{fig:interpolation-parameter-vs-guidance-scale}
\end{figure}

\begin{figure}
    \centering
    \includegraphics[width=\linewidth]{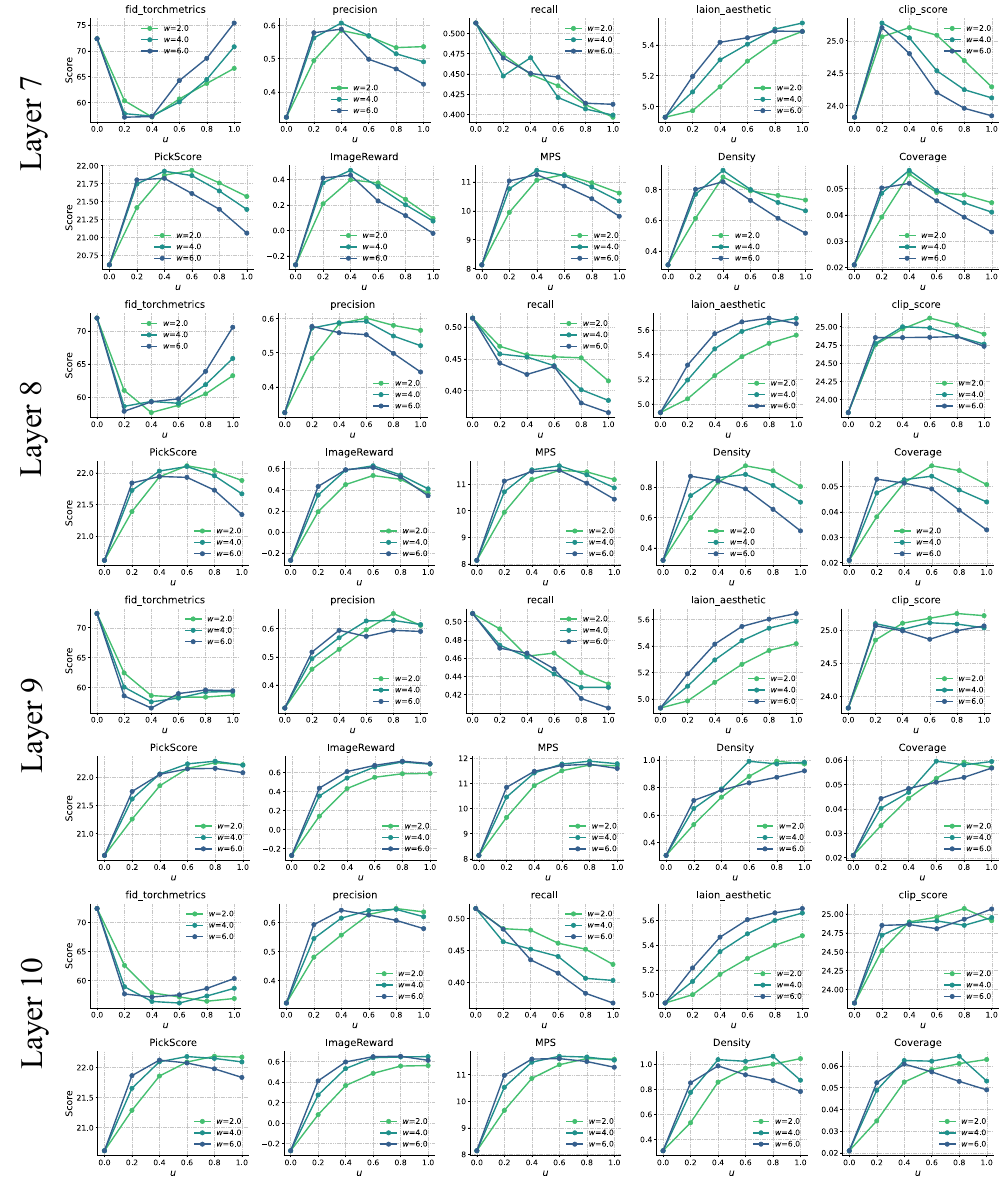}
    \caption{\textbf{Quantitative results of interpolation in SoftPAG, on layer $7, 8, 9, 10$ with different metrics.}}
    \label{fig:softPAG_inter}
\end{figure}

\begin{figure}
    \centering
    \includegraphics[width=\linewidth]{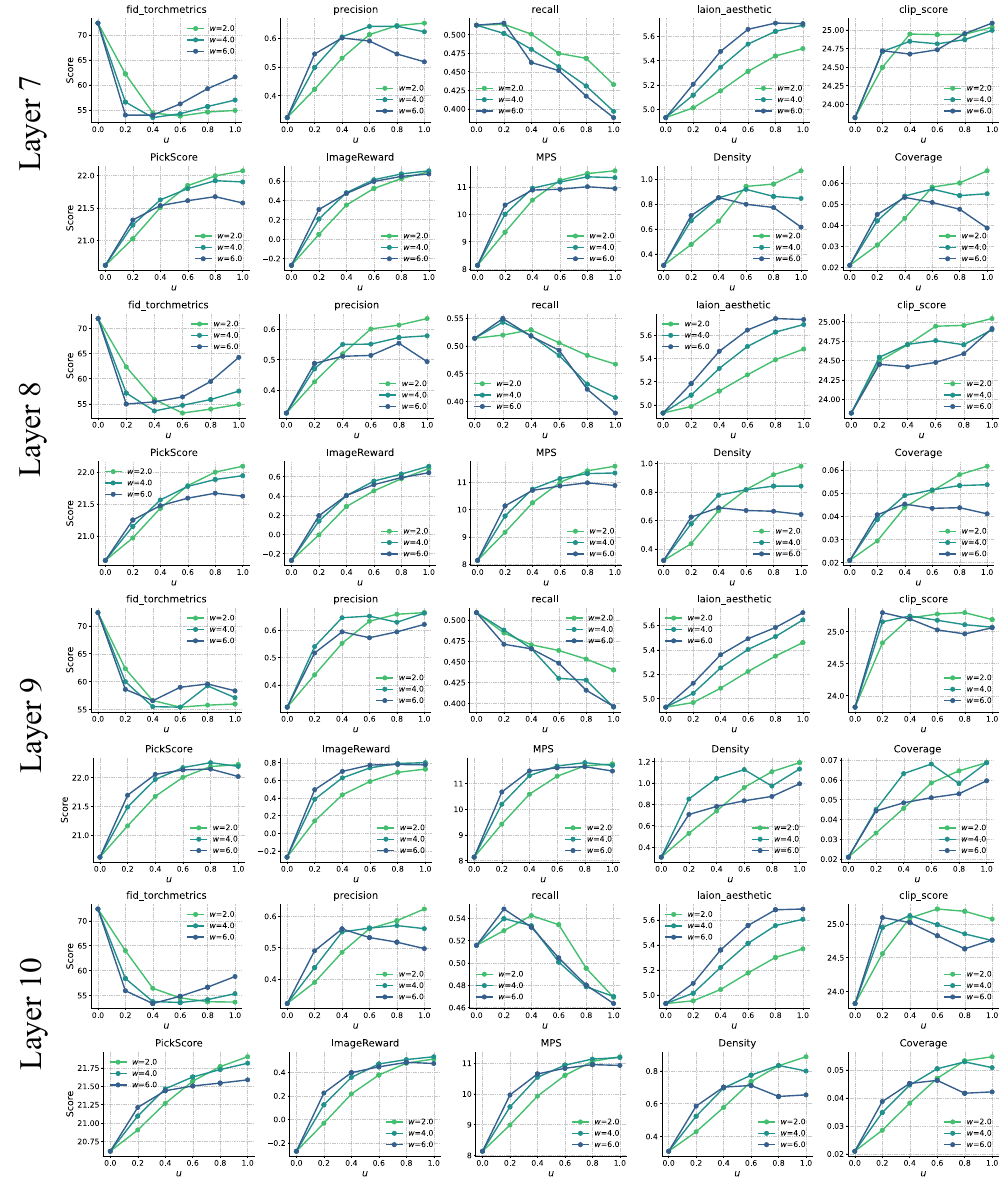}
    \caption{\textbf{Quantitative results of interpolation in UG, on layer $7, 8, 9, 10$ with different metrics.}}
    \label{fig:UG_inter}
\end{figure}

We measured the quality of images generated by layers 7, 8, 9, and 10, which produce the highest-quality outputs. Fig.~\ref{fig:softPAG_inter} depicts the results of SoftPAG, which interpolates between the identity matrix and the attention maps, while Fig.~\ref{fig:UG_inter} depicts the results of UG, which interpolates between the uniform matrix and the attention maps.


\clearpage


\phantomsection

\clearpage
\section{HeadHunter: Implementation and additional analysis}
\label{sup:sec:headhunter}

\subsection{Implementation details}

\label{sup:subsec:headhunter-impl-detail}

\paragraph{Improving general image quality.}

To run HeadHunter, we used 20 prompts ($M = 20$), with $k = 24$ and $R = 1$. All prompts and the corresponding seeds are listed in Tab.~\ref{tab:general_prompt_list}. Since SD3 contains 24 layers with 24 attention heads per layer, we consider the full set of $N = 24 \times 24 = 576$ attention heads. For head-level guidance, we set the guidance scale to 3.0 and the number of inference steps to 20. Additionally, we set the interpolation parameter $u = 0.25$ in Eq.~\ref{eq:identity-interpolation} to avoid excessive over-smoothing effects.

While it is possible to set $R > 1$ to run HeadHunter iteratively, we found that a single round suffices in practice. This is based on the compositional nature of head-level guidance. Heads that individually improve quality tend to combine well, yielding consistent improvements in overall generation quality. As shown in Fig.~\ref{fig:head-numbers}, even a compact set of heads selected in a single round can outperform the best-performing layer-level guidance.


\paragraph{Improving style-oriented quality.}
For SD3, we evaluated a total of 55 style prompts for both quantitative and qualitative analysis. Since SD3 consists of 24 layers with 24 attention heads per layer, we considered all $N = 24 \times 24 = 576$ attention heads. For head-level guidance, we set the guidance scale to 3.0 and used 20 inference steps.

For FLUX.1-Dev, we used a total of 23 style prompts. The model comprises 57 layers with 24 attention heads per layer, resulting in $N = 57 \times 24 = 1368$ attention heads. Head-level guidance was applied with a guidance scale of 8.0 and 15 inference steps.

In both cases, we used 5 content prompts ($M = 5$) per style. The full list of content prompts is provided in Tab.~\ref{tab:content_prompt_list}.

\subsection{Additional results and analysis}
\label{sup:subsec:headhunter-impl-additional-restuls}

\subsubsection{Additional results}
We present additional qualitative results for style-oriented quality improvement using HeadHunter in \Cref{fig:qual_sd3_sunlit_spring,fig:qual_sd3_line_flat,fig:qual_sd3_cinematic_psychedelic}. Each row corresponds to the result obtained by perturbing the head set selected at round $r$ of Alg.\ref{alg:headhunter-iter}. The top row shows unguided generation, and the lower rows show results with an increasing number of selected heads applied. As more heads are progressively added, the generated images align better with the target style while maintaining visual plausibility.

\begin{figure}[!t]
    \centering
    \includegraphics[width=\linewidth]{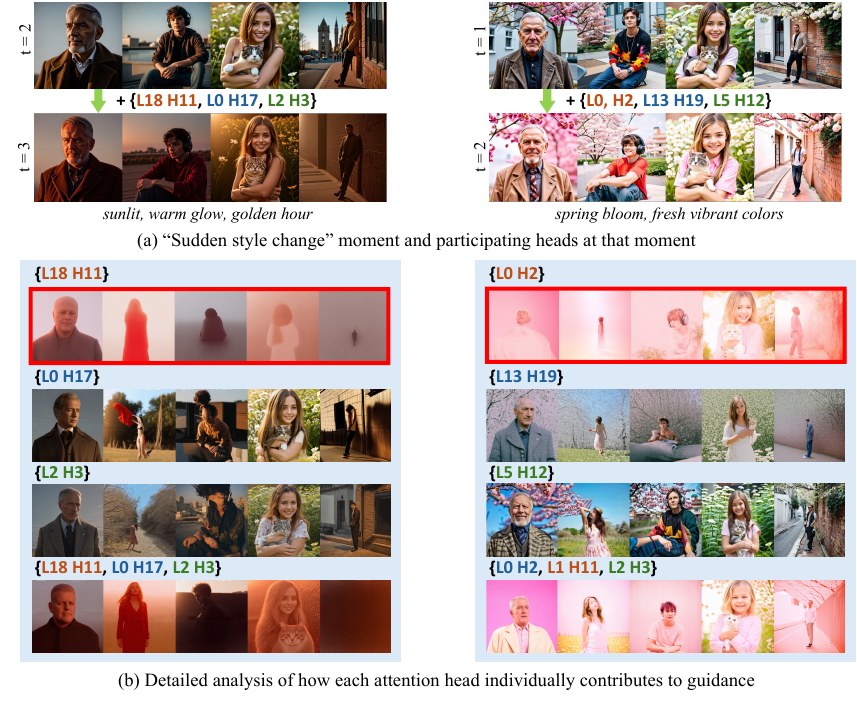}
    \caption{\textbf{Role of individually weak heads.} For sudden stylistic transition moments (e.g., adding global warmth or pink hues as shown in (a)), we visualize the effect of newly added heads in (b). One of these heads generate blurry reddish or pinkish outputs when used alone (see red box). Although they fail to produce meaningful content when used independently, they contribute effectively when composed with previously selected structural heads. This highlights the importance of iterative strategy since those heads are unlikely to be selected by one-shot evaluation due to low quality.}
    \label{fig:suboptimal_full}
    \vspace{-20pt}
\end{figure}

For example, for the style prompt \textit{``sunlit, warm glow, golden hour''}, we observe an increasing glow effect, culminating in a global golden-hour tone from the fourth row onward. We provide a more detailed analysis of the contributing heads in Sec.~\ref{sup:para:suboptimal-heads} and Fig.~\ref{fig:suboptimal},~\ref{fig:suboptimal_full}.

Interestingly, the framework is effective not only for photorealistic styles such as \textit{``cinematic lighting''}, but also for abstract or simplified styles like \textit{``line art''} or \textit{``flat paper cut''}. Since this style adaptation is achieved without modifying any model parameters, it enables a reusable and lightweight preference tuning.

\begin{figure}[!p]
    \centering
    \includegraphics[]{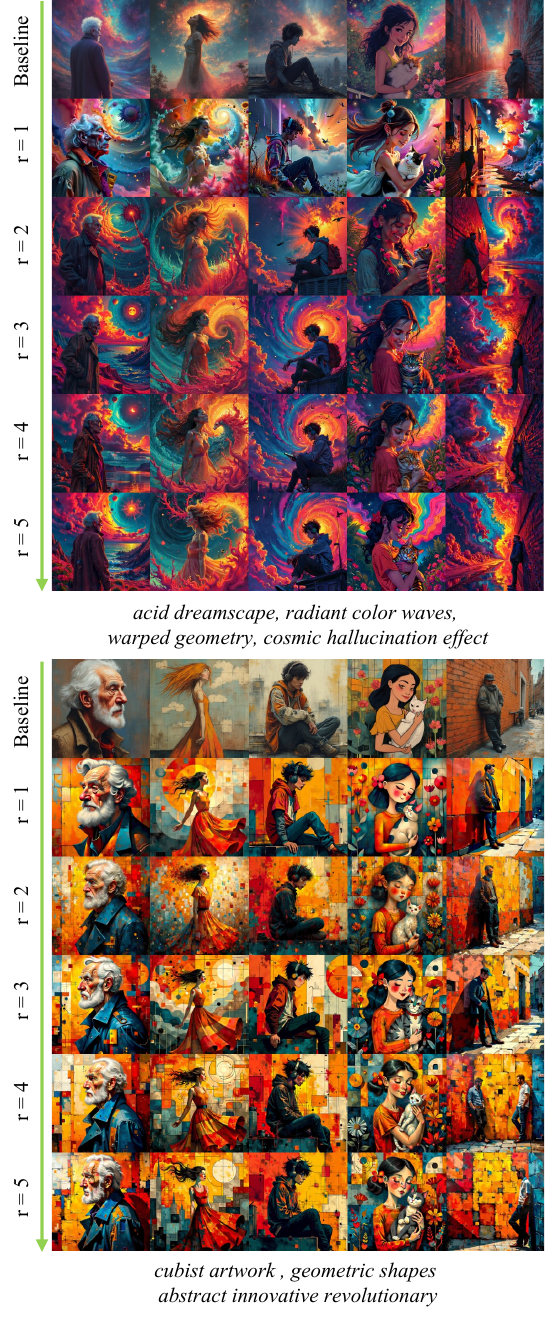}
    \caption{\textbf{Qualitative results of HeadHunter for style-oriented image quality improvement on FLUX.1-Dev.}}
    \label{fig:qual_flux_acid_cubist}
\end{figure}

\begin{figure}[!p]
    \centering
    \includegraphics[]{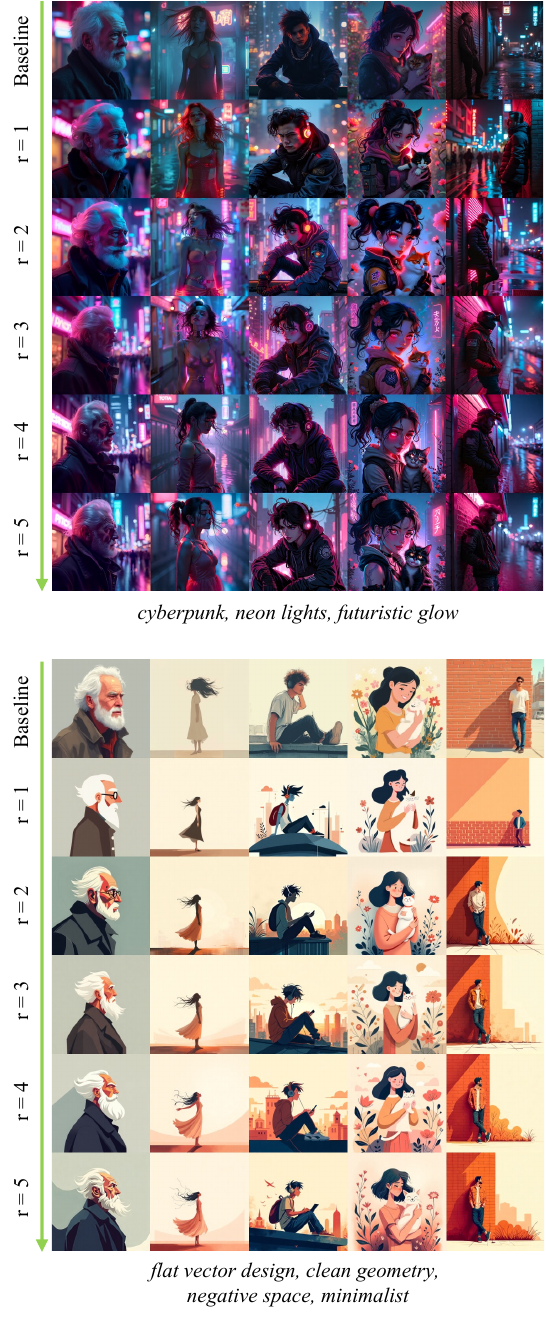}
    \caption{\textbf{Qualitative results of HeadHunter for style-oriented image quality improvement on FLUX.1-Dev.}}
    \label{fig:qual_flux_cyberpunk_flat}
\end{figure}

\begin{figure}[!p]
    \centering
    \includegraphics[]{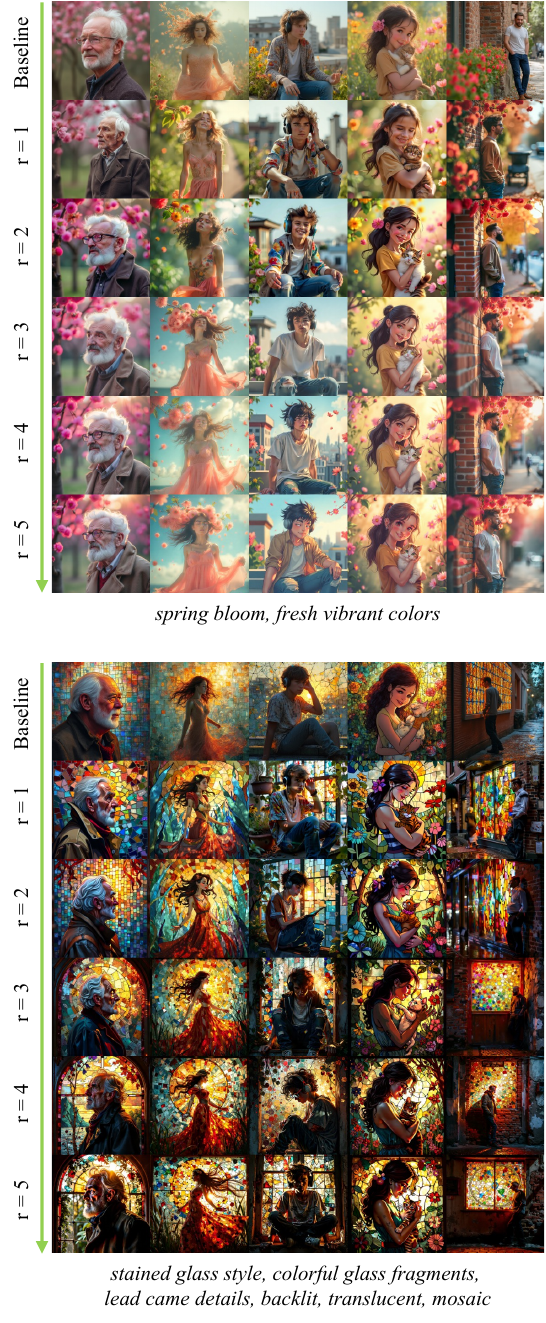}
    \caption{\textbf{Qualitative results of HeadHunter for style-oriented image quality improvement on FLUX.1-Dev.}}
    \label{fig:qual_flux_spring_stained}
\end{figure}

HeadHunter can benefit from with model size. Since larger models have more expressive attention heads, they can be more effectively leveraged in head-level perturbation guidance. In Fig.~\ref{fig:qual_sd3_sunlit_spring}-\ref{fig:qual_flux_spring_stained}, we provide qualitative results of style-oriented quality improvement conducted on FLUX.1-dev~\cite{blackforestlabs_flux1_dev}. 

\subsubsection{Additional analysis}
\begin{figure}[!t]
    \centering
    \includegraphics[width=\linewidth]{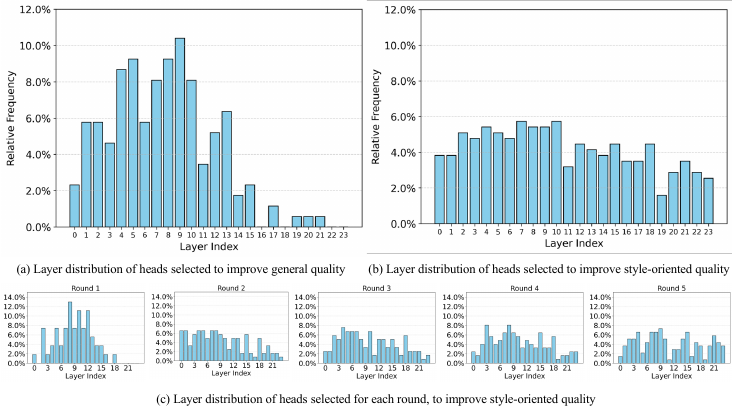}
    \caption{\textbf{Layer distribution of heads selected by HeadHunter for different objectives.} Effective heads are not confined to specific layers and the distribution is different with each objective, highlighting the limitations of layer‐level guidance and the necessity of objective‐specific head selection.  
(a) In general image quality setting, selections concentrate in early–mid layers (2.9\% in layers \(\geq16\)).  
(b) In style‐oriented quality setting, selections spread broadly, with 24.8\% in layers \(\geq16\).  
(c) Round-wise distributions of (b) begin with an early-layer bias similar to (a), but gradually flatten over rounds, mirroring (b). This trend aligns with qualitative observations. Round 1 primarily improves general quality, whereas stylization becomes more prominent in subsequent rounds.}
    \label{fig:head_distribution}
    \vspace{-10pt}
\end{figure}

\paragraph{Distribution of selected heads across layers.}
We investigates the layer distribution of selected attention heads within a model's architecture, specifically examining whether impactful heads are localized to particular layers and if their distribution varies by objective. For the general quality setting, we select the top-15 heads per prompt and aggregate them, excluding duplicates (Fig.~\ref{fig:head_distribution} (a)). For style-oriented quality setting, we repeat the process per style (Fig.~\ref{fig:head_distribution} (b)).

Our findings reveal that effective heads are not concentrated in any single layer. Even the most frequently selected layer accounts for less than 12\% of all chosen heads across both objectives (Fig.~\ref{fig:head_distribution} (a), (b)). This dispersed distribution suggests that a layer-wise guidance approach, which uniformly perturbs all heads within a layer, may be suboptimal. Such an approach risks overlooking critical heads in other layers while simultaneously applying modifications to irrelevant heads.

Furthermore, we observed distinct distribution patterns dependent on the objective. Heads identified for improving general image quality predominantly reside in early to mid-layers, with a mere 2.9\% located in layers $\geq$16. In contrast, heads contributing to style-oriented quality improvement exhibit a significantly broader distribution, with 24.8\% found in deeper layers ($\geq$16).

\begin{figure}[!p]
    \centering
    \includegraphics[]{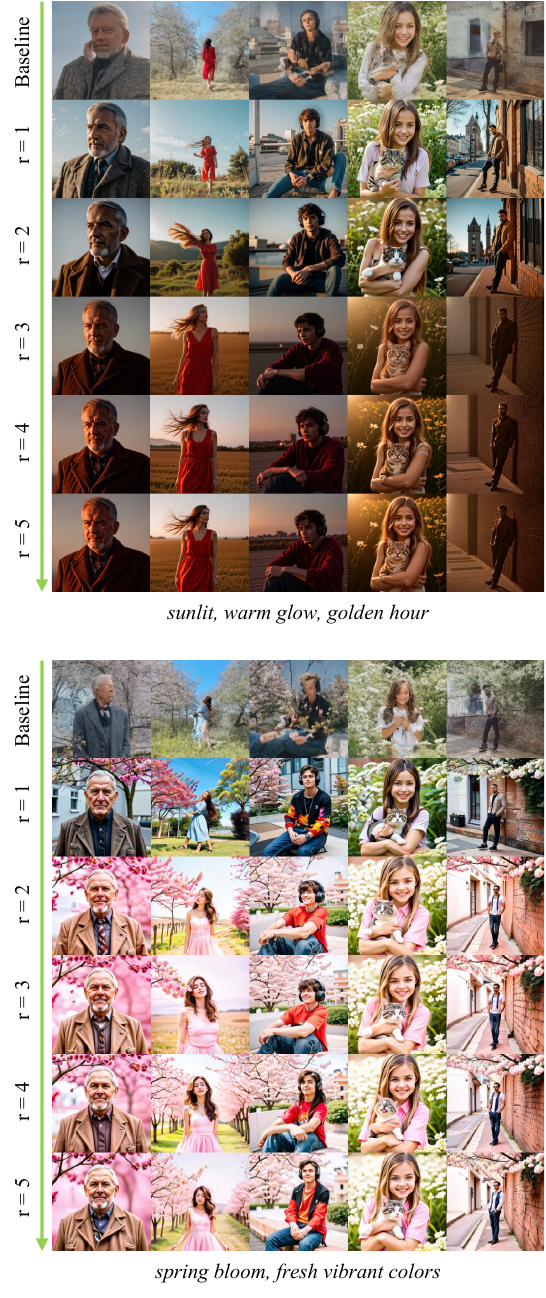}
    \caption{\textbf{Qualitative results of HeadHunter for style-oriented image quality improvement on SD3.}}
    \label{fig:qual_sd3_sunlit_spring}
\end{figure}

\begin{figure}[!p]
    \centering
    \includegraphics[]{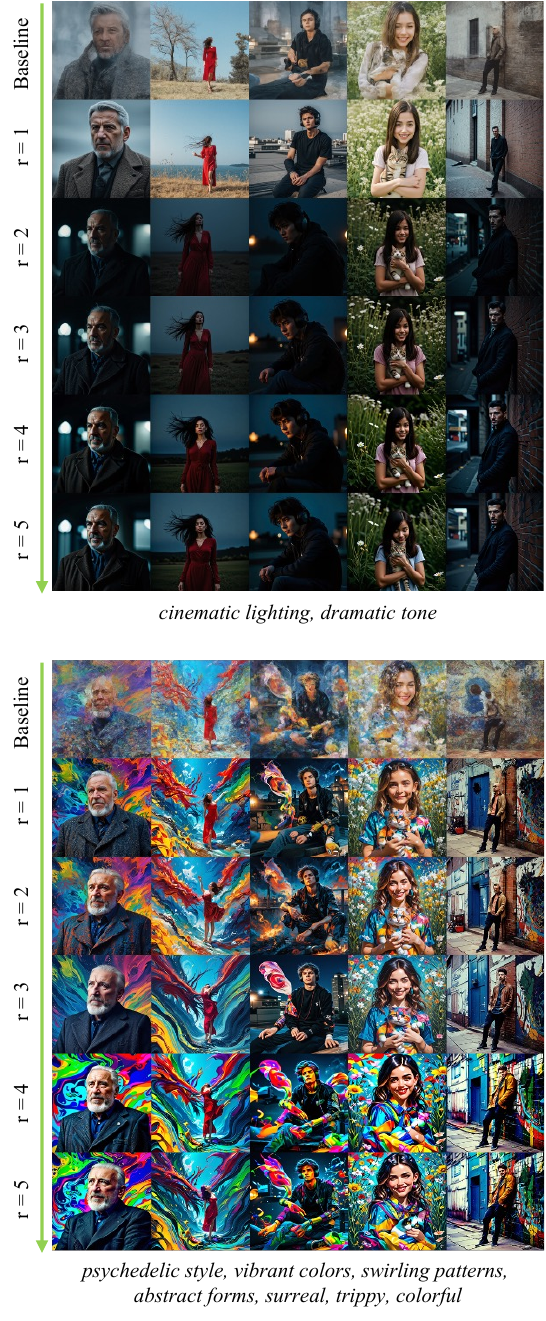}
    \caption{\textbf{Qualitative results of HeadHunter for style-oriented image quality improvement on SD3.}}
    \label{fig:qual_sd3_cinematic_psychedelic}
\end{figure}

\begin{figure}[!p]
    \centering
    \includegraphics[]{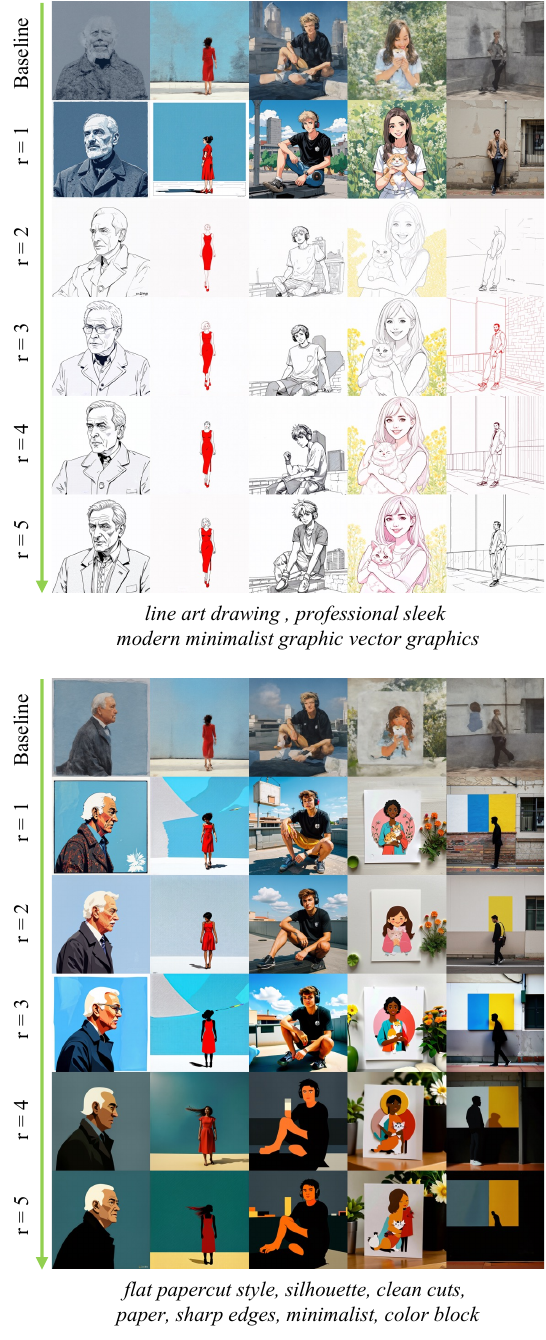}
    \caption{\textbf{Qualitative results of HeadHunter for style-oriented image quality improvement on SD3.}}
    \label{fig:qual_sd3_line_flat}
\end{figure}

We further examine the evolution of head distribution during HeadHunter's iterative refinement for style-oriented quality setting in Fig.~\ref{fig:head_distribution} (c)). Initially, selected heads are concentrated in mid-layers, mirroring the pattern observed for general quality setting. However, as rounds progress, the distribution become notably flatter, indicating a broader engagement of heads across layers for stylistic refinement. This quantitative shift aligns with qualitative observations.As shown in Fig.~\ref{fig:qual_sd3_sunlit_spring} (spring bloom, fresh vibrant colors), Fig.~\ref{fig:qual_sd3_cinematic_psychedelic} (cinematic lighting, dramatic tone), and Fig.~\ref{fig:qual_sd3_line_flat} (line art drawing ...), the second row (Round 1) shows improved image quality but little stylistic traits. However, from round 2 onward, the style becomes much more pronounced..

In conclusion, our research indicates that (1) effective heads are not localized to specific layers, thereby challenging the efficacy of layer-level guidance, and (2) head distributions are objective-dependent, underscoring the necessity of objective-specific selection. These findings collectively emphasize the critical role of targeted head-retrieval and guidance frameworks, such as HeadHunter, in fully harnessing the diverse functionalities of attention mechanisms.

\paragraph{Does the style enhancement occur even without style prompts?}

In our style-oriented quality improvement experiments, we search for heads that consistently enhance a given style by using composite prompts of the form ``\texttt{style, content}''. Then a natural question arises: \textit{Do these heads still express their stylistic traits when the style is not explicitly specified in the prompt or even under unconditional generation?}

\begin{figure}[!t]
    \centering
    \includegraphics[width=\linewidth]{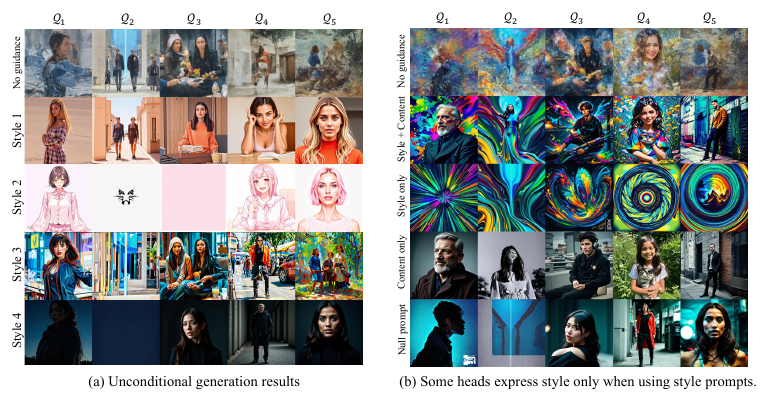}
\caption{\textbf{Investigation whether style enhancement occurs without style prompt when applying HeadHunter (style-oriented quality setting).} $\mathcal{Q}$ indicates prompt and seed pairs. (a) In some cases, applying HeadHunter in even an unconditional setting amplifies the corresponding style to some extent. Style 1: ``art nouveau style, elegant, decorative, curvilinear forms, nature-inspired, ornate, detailed'', Style 2: ``line art drawing, professional sleek modern minimalist graphic vector graphics", Style 3: ``cubist abstraction, fragmented planes, bold geometric angles'', Style 4: ``cinematic lighting, dramatic tone''. (b) In the other cases, style enhancement depends on the presence of style prompts. For these heads, the stylization effect weakens without explicit style descriptions, highlighting the importance of including style prompts during the head selection process. Style prompt : "psychedelic style, vibrant colors, swirling patterns, abstract forms, surreal, trippy, colorful
reflective voids, perceptual disorientation, sinuous forms, color psychology"}
    \label{fig:uncond-style-failure}
    \vspace{-10pt}
\end{figure}

To explore this, we apply the heads selected by HeadHunter in an unconditional setting (i.e., with a null prompt, ``''). As shown in Fig.~\ref{fig:uncond-style-failure} (a), we find that even with a null prompt, using these heads can still induce visual traits that align with the original style to some extent. This suggests that certain heads may encode intrinsic stylistic properties that influence generation regardless of prompt or initial noise.

Yet, this effect is not guaranteed across all styles. As can be seen in Fig.~\ref{fig:uncond-style-failure} (b), in some cases, while style traits are strongly expressed with style prompts, they become weak or ambiguous when only content prompts are used or under unconditional generation. This may be due to polysemanticity in head behavior. This emerging style consistency when using style prompts indirectly supports the validity of including style prompts during HeadHunter’s search process in style-oriented quality setting.

\paragraph{Computational efficiency.}
HeadHunter requires a one-time search process to identify effective attention heads for a given objective. If we denote the number of prompt–seed pairs as $M$, total attention heads as $N$, number of rounds as $R$, and the costs for generation and evaluation as $C_{\text{gen}}$ and $C_{\text{eval}}$, the overall cost is approximately $MN(C_{\text{gen}} + C_{\text{eval}})R$. In our experiments using \texttt{stable-diffusion-3-medium} with 20-step Euler sampling and PickScore~\cite{kirstain2023pick} as the objective, this process takes roughly 3 hours on 8$\times$NVIDIA H100 GPUs.

However, this cost is amortized over repeated use. Once a set of heads is selected for a given model and configuration, it can be reused across different prompts and latent seeds—unlike test-time scaling approaches~\cite{ma2025inference,li2025reflect} that require per-sample optimization. In the previous section, we further demonstrate that head sets found using a small subset of prompts generalize well to broader and more diverse prompt distributions. At inference time, attention perturbation guidance adds negligible overhead, as it simply modifies a fixed subset of attention maps without altering the model architecture or requiring additional forward passes.


\FloatBarrier


\clearpage
\section{Additional results and analysis of head-level perturbation guidance}
\label{sup:sec:additional_heads_analysis}
In this section, we provide an extended analysis of head-level perturbation guidance with a focus on interpretable visual concepts. Specifically, Sec.~\ref{sup:subsec:all-heads} presents more individual head-level guidance results in Stable Diffusion 3 and FLUX.1-Dev. In Sec.~\ref{sup:subsec:interpretable_heads}, we showcase examples of head-level guidance that exhibit clearly interpretable semantic effects. Finally, Sec.~\ref{sup:subsec:more_combinations} explores the compositional behavior of head-level guidance and how combining them yields richer visual outcomes.

\clearpage
\subsection{Additional results of individual head-level guidance}
\label{sup:subsec:all-heads}

\begin{figure}[!h]
    \centering
    \includegraphics[width=0.98\linewidth]{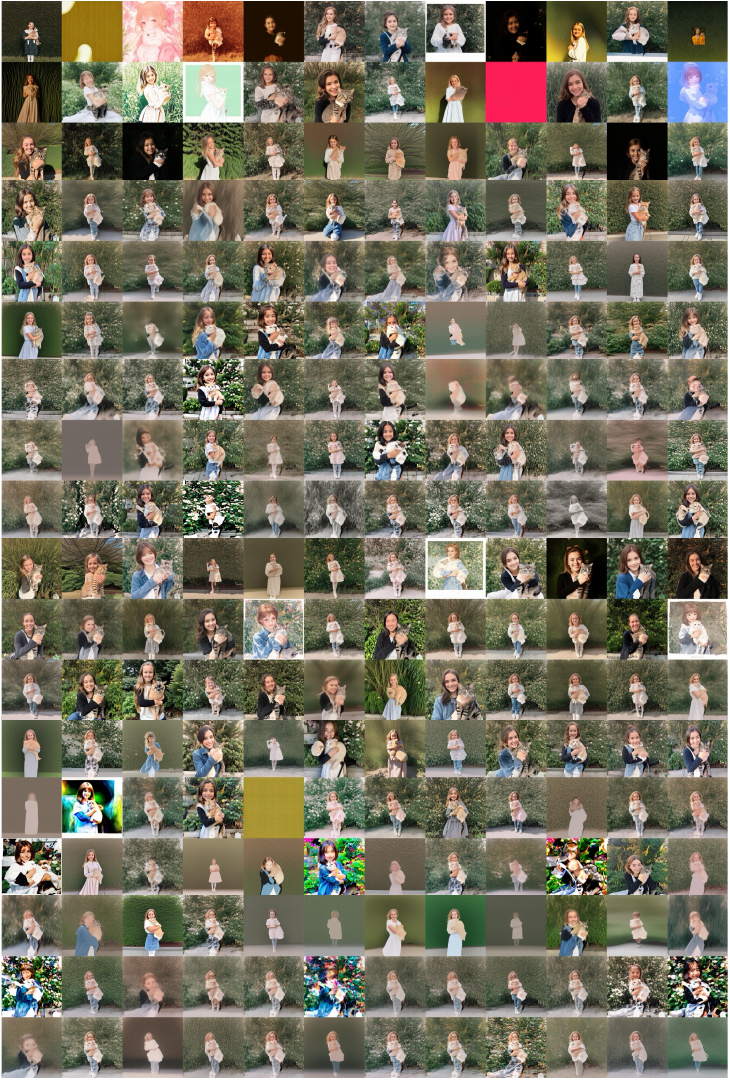}
    \caption{\textbf{Individual head-level guidance results on SD3.} Each cell corresponds to the output guided by perturbing a single head. Some heads induce notable effects, such as changes in lighting, structure, or color.}
    \label{fig:all-heads-sd3}
\end{figure}

\begin{figure}[!p]
    \centering
    \includegraphics[width=\linewidth]{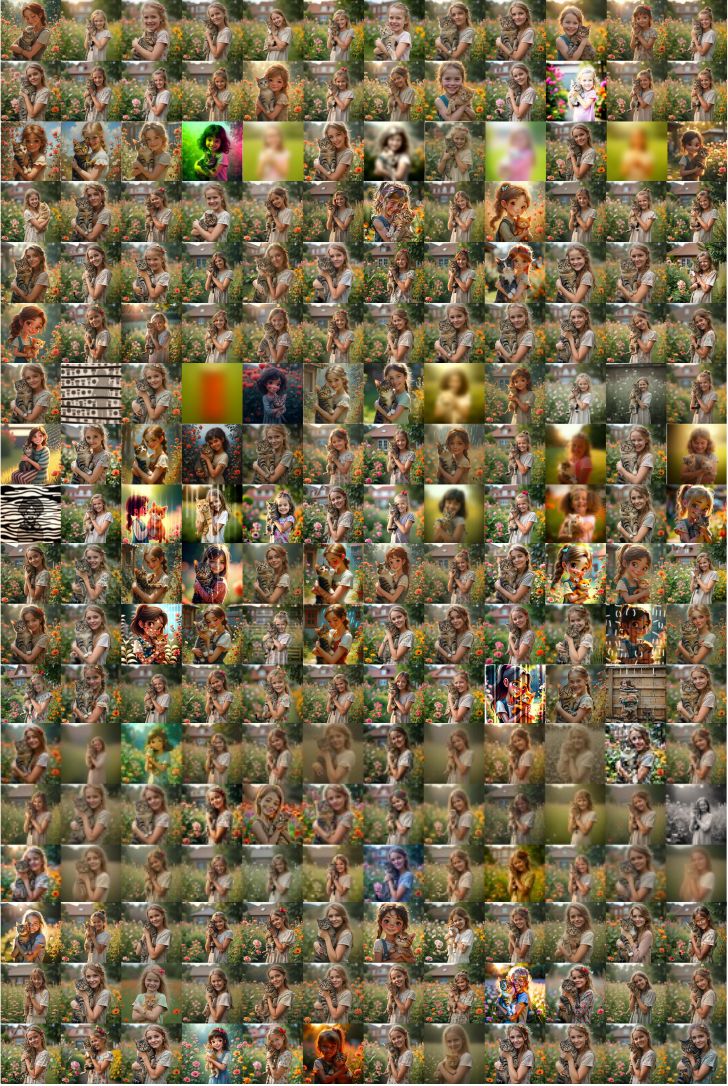}
    \caption{\textbf{Individual head-level guidance results on FLUX.1-Dev.} Each cell corresponds to the output guided by perturbing a single head. Some heads induce notable effects, such as changes in lighting, structure, or color.}
    \label{fig:all-heads-flux}
\end{figure}

\clearpage
\subsection{Additional results on interpretable head-level guidance}
\label{sup:subsec:interpretable_heads}

\begin{figure}[!h]
    \centering
    \includegraphics[width=\linewidth]{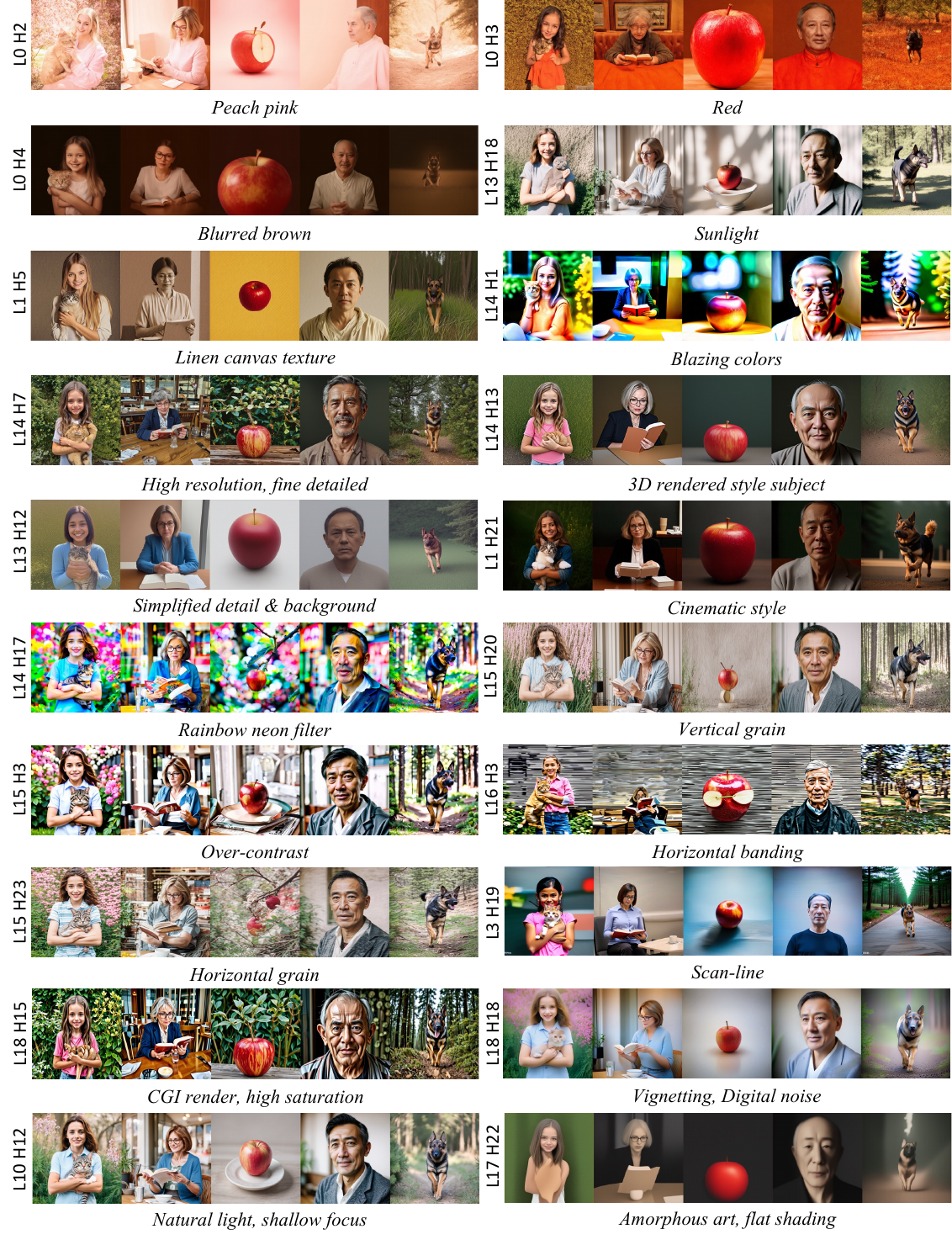}
    \caption{\textbf{Interpretable heads in SD3.} Some heads exhibit consistent and clear visual effects when used for guidance. For instance, heads that add glow, alter geometry, or introduce specific lighting styles. These interpretable heads support the view that attention heads encode specific semantic concepts.}
    \label{fig:interpretable_sd3}
\end{figure}

\begin{figure}[!h]
    \centering
    \includegraphics[width=\linewidth]{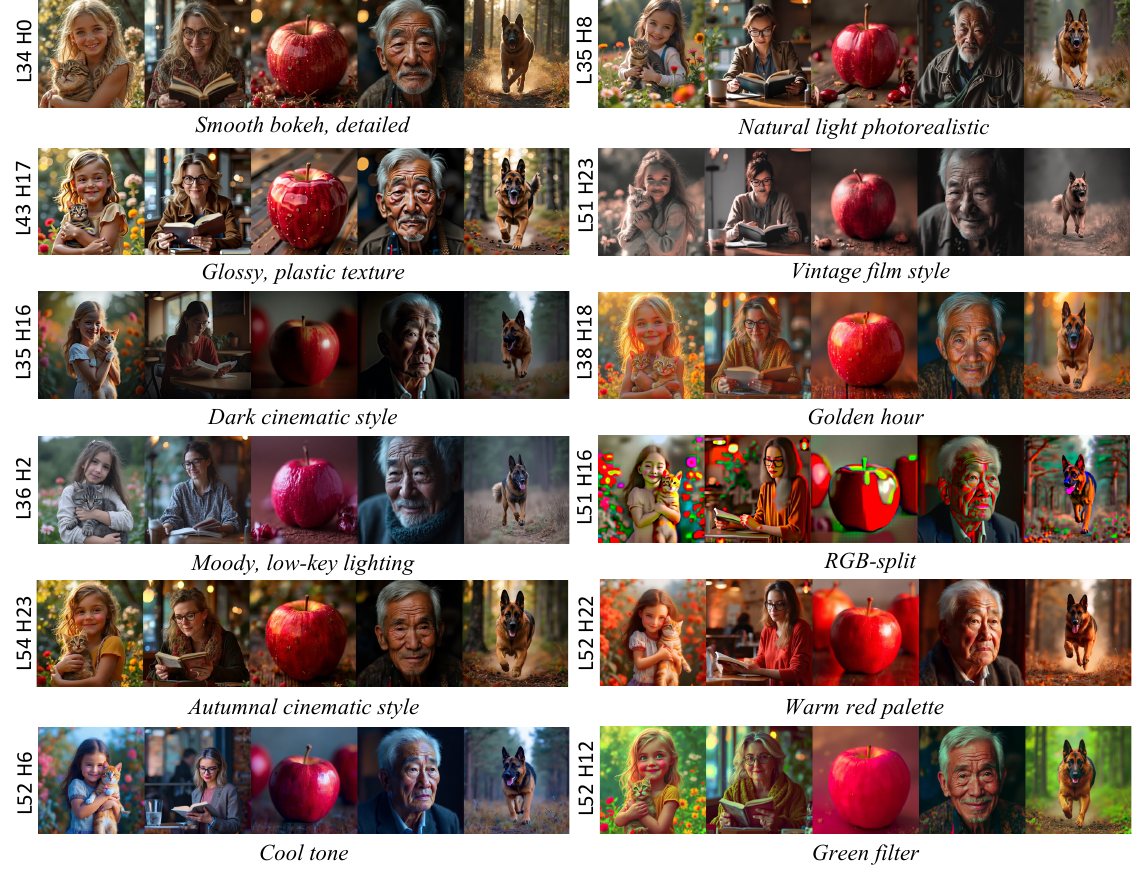}
    \caption{\textbf{Interpretable heads in FLUX.1-Dev.} Similar to SD3, FLUX also contains heads with distinctive effects, suggesting the generality of head-level interpretability across architectures.}
    \label{fig:interpretable_flux}
\end{figure}

\FloatBarrier

\clearpage
\subsection{Additional results on the combinational effect of head-level guidance}
\label{sup:subsec:more_combinations}

\begin{figure}[!h]
    \centering
    \includegraphics[width=\linewidth]{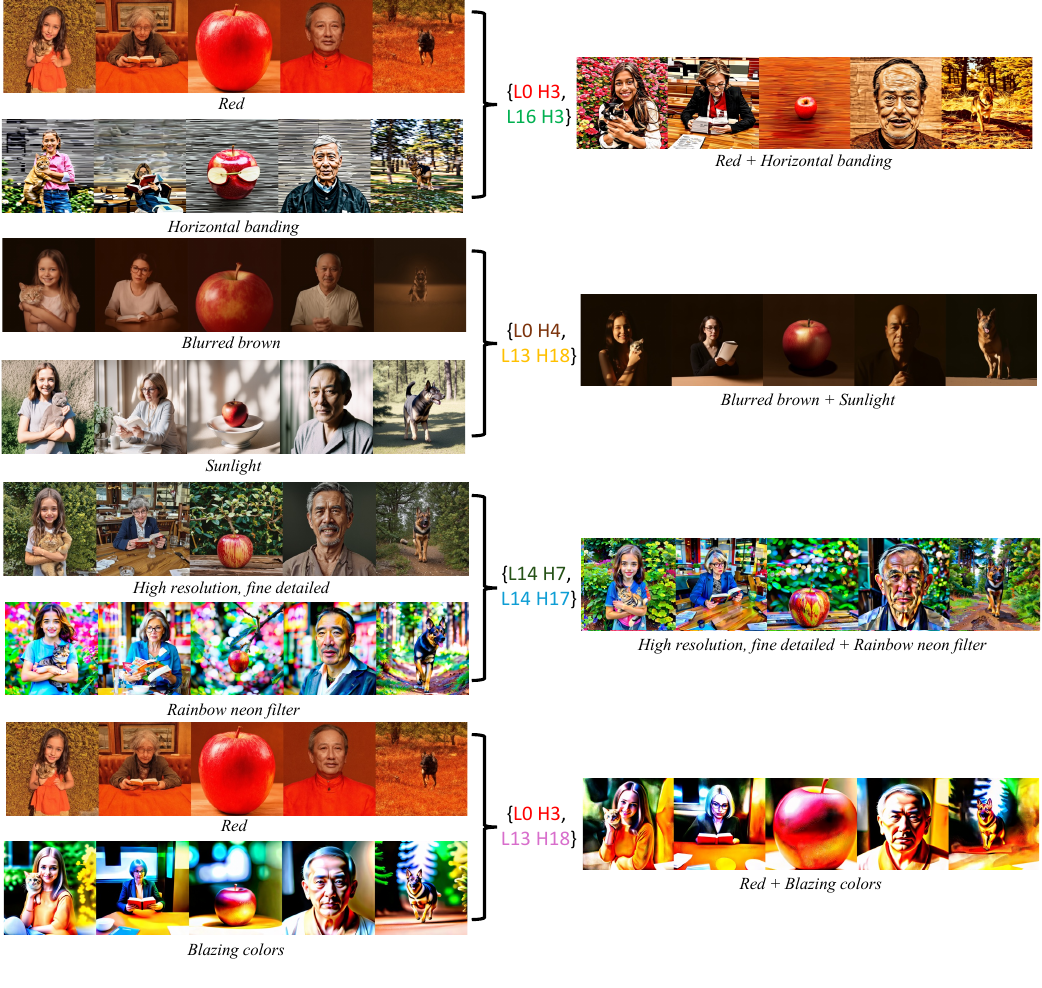}
    \caption{\textbf{Compositional effects of head-level guidance in SD3.} Combining heads leads to enriched outputs by \textit{blending} their individual effects. This shows that attention heads can be composed to control more complex or stylized generations.}
    \label{fig:more-combination}
\end{figure}

\begin{figure}[!h]
    \centering
    \includegraphics[width=\linewidth]{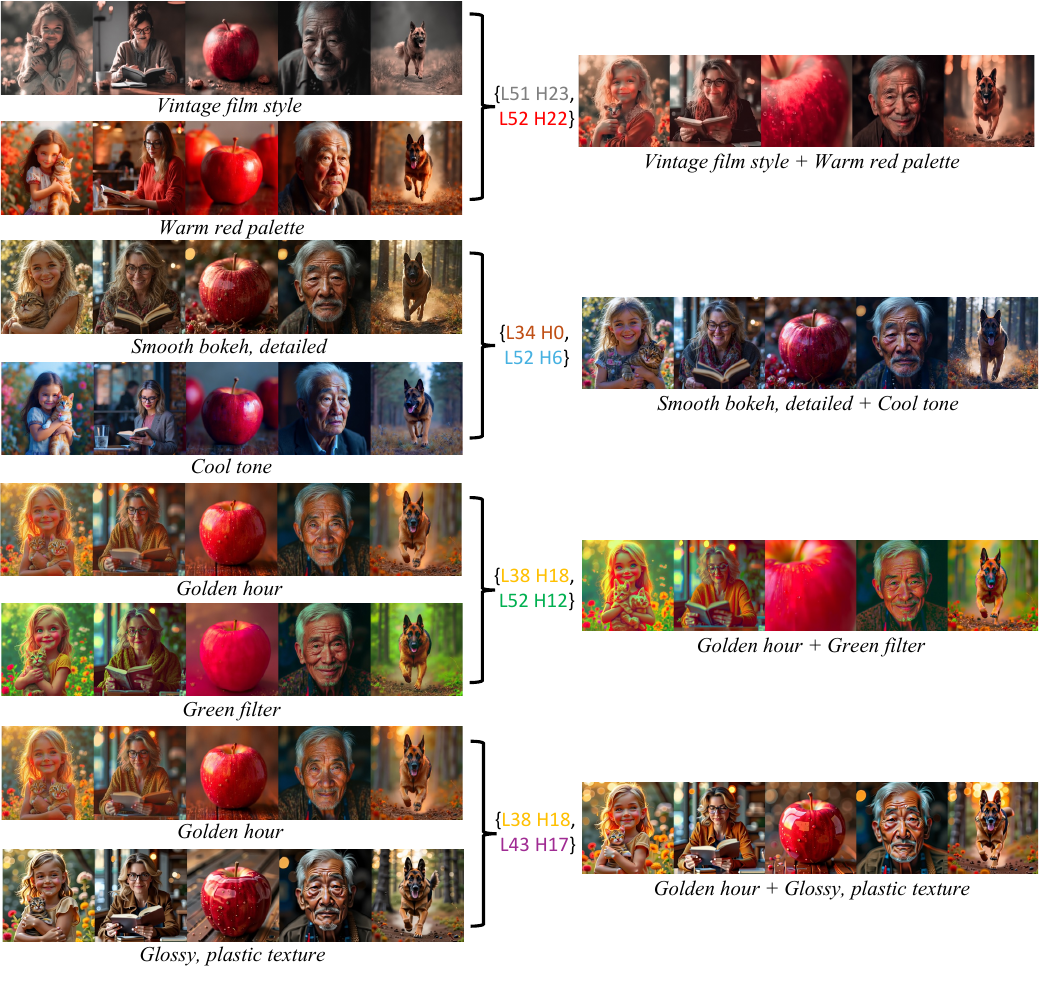}
    \caption{\textbf{Compositional effects of head-level guidance in FLUX.1-Dev.} Similar head-level guidance effects can be composed in FLUX.1-Dev to amplify or adjust stylistic elements.}
    \label{fig:concept_combination_flux}
\end{figure}

\begin{figure}[!h]
    \centering
    \includegraphics[width=\linewidth]{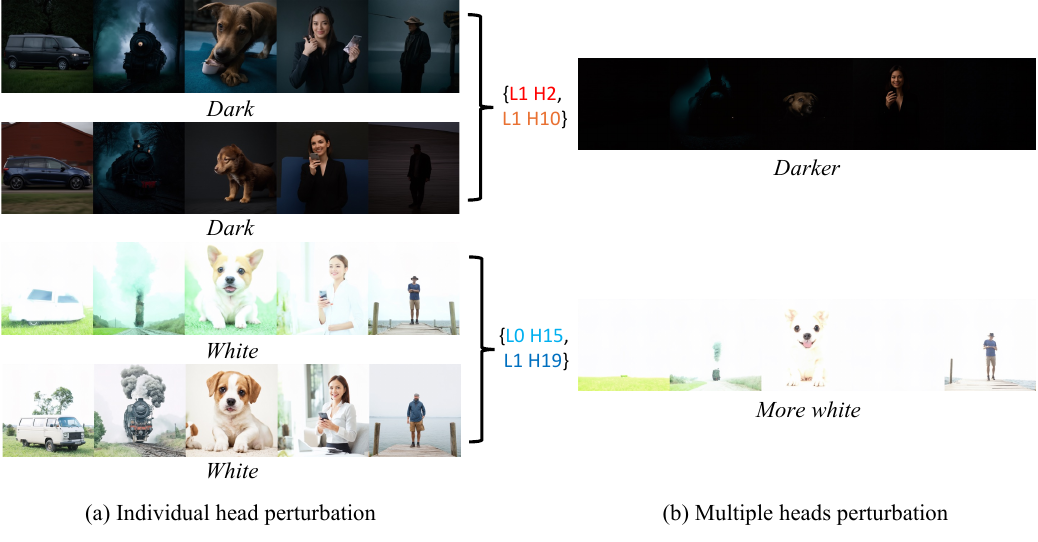}
    \caption{\textbf{Concept amplification by combining stylistic heads.} Composing ``dark heads'' makes the generation darker, while ``white heads'' make it brighter. This demonstrates fine-grained control over stylistic dimensions via head-level selection.}
    \label{fig:dark-white}
\end{figure}

\section{Head-level perturbation guidance with different perturbation methods}
\label{sup:head-perturbation-using-other-perturbations}

\begin{figure}[h]
    \centering
    \includegraphics[width=\linewidth]{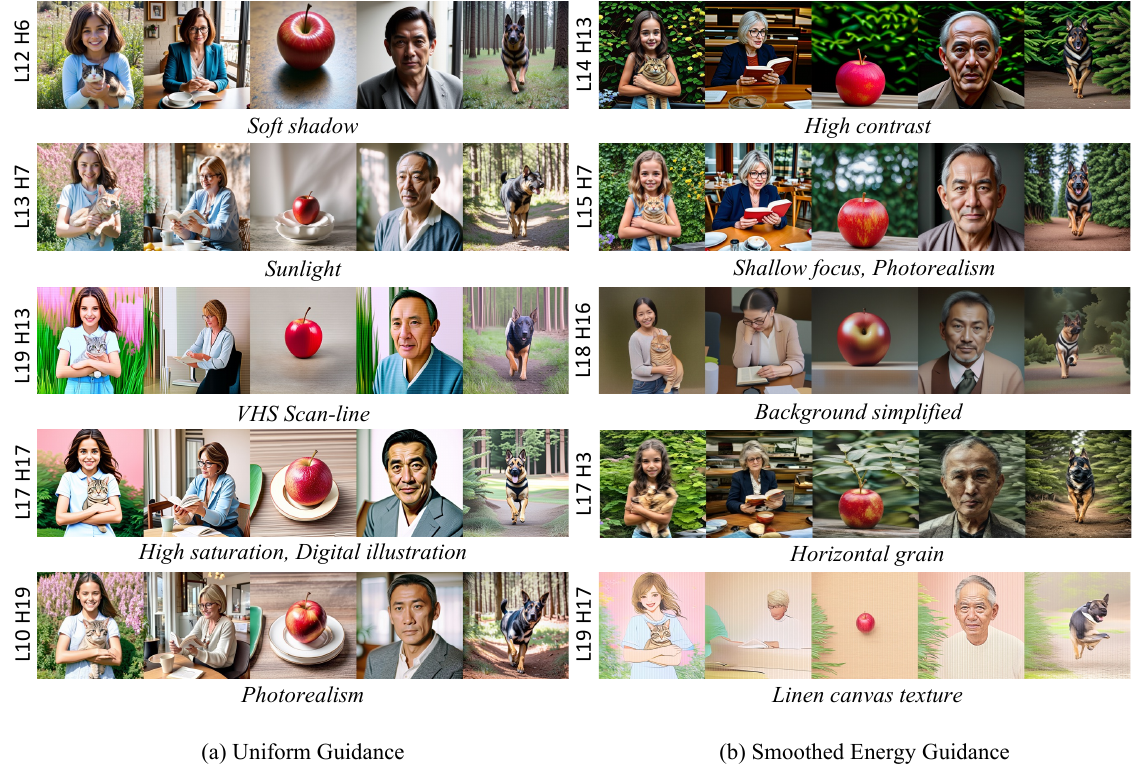}
    \caption{\textbf{Head-Level perturbation guidance with different perturbations.} We show interpretable heads with different perturbation methods.}
    \label{fig:interpretable-different-perturbations}
\end{figure}

In the main paper, we show the head-level analysis mostly on identity-matrix replacement perturbation. In Fig.~\ref{fig:interpretable-different-perturbations}. We show the results with another perturbation and the analysis. Note that one can freely choose any other perturbation methods, and we show some cases as examples.

\label{sup:sec:uniform_head_analysis}

\clearpage
\section{Experimental details}
\label{sup:sec:experimental-details}



Unless otherwise specified, all images are generated using \texttt{Stable Diffusion 3-Medium} with 20 Euler sampling steps. The default guidance scale is set to $w = 5.0$.

\paragraph{Fig.~\ref{fig:motivation}}  
\textbf{Motivating examples for head-level guidance.}  
Perturbations are applied individually to heads from layers 1, 3, and 14. Prompts are sampled from the MS-COCO validation set.

\paragraph{Fig.~\ref{fig:head-per-layer}}  
\textbf{Generated images from head- and layer-level perturbation guidance.}  
In (a), heads with PickScore above a fixed threshold are selected (indicated by red boxes). Prompts include \texttt{“a medieval castle in the forest”} and \texttt{“a futuristic cityscape”}.

\paragraph{Fig.~\ref{fig:special-heads}}  
\textbf{Effect of head-level guidance on concept amplification.}  
Each head is perturbed independently. Prompts used include: \texttt{"smiling girl holding a cat, in a flower garden"}, \texttt{"middle-aged woman with glasses reading a book in a café"},
\texttt{"A macro shot of a red apple"},
\texttt{"a close up portrait of an elderly Japanese man, soft lighting, 8k"}, and \texttt{"A German shepherd bounding through a pine forest"}. Guidance scale is set to $w = 5.0$.

\paragraph{Fig.~\ref{fig:head-numbers}}  
\textbf{Results of HeadHunter for general image quality improvement.}  
HeadHunter is applied with $R = 1$ and $k = 24$. Prompts used include \texttt{“smiling girl holding a cat, in a flower garden”}, \texttt{“businessman with sleek, glass-shiny black hair neatly parted, waiting for a morning train on a city platform”}, and \texttt{“towering lighthouse casting a rotating beam across storm-tossed waves at twilight”}. Metrics such as PickScore and AES are used for evaluation.

\paragraph{Fig.~\ref{fig:AI_interpolation}}  
\textbf{Linear interpolation between attention map $\attnmap$ and identity matrix $\mathbf{I}$ (SoftPAG).}  
Images are generated with perturbation applied to layer 9. Prompts are generated from ChatGPT.

\paragraph{Fig.~\ref{fig:AU_interpolation}}  
\textbf{Linear interpolation between attention map $\attnmap$ and uniform matrix $\mathbf{U}$ (UG).}  
Perturbation is applied to layer 9. Prompts are the same as those used in Fig.~\ref{fig:AI_interpolation}.


\section{Experimental resources}
    All experiments, including testing perturbation methods, analyzing head-level guidance, running HeadHunter, and conducting ablation studies, are performed using a mix of 8 NVIDIA H100 GPUs, 6 NVIDIA RTX 3090 GPUs, and 2 NVIDIA A6000 GPUs.

\section{Limitations \& Broader Impacts}
While our method is effective, there remains room for improving its efficiency. Running HeadHunter iteratively can be computationally intensive, especially for large-scale models with many heads. That said, the selected heads tend to generalize well across prompts and latents, making reuse practical. Exploring faster head and parameter selection strategies offers an exciting direction for future work.

Our work improves quality and stylistic control in diffusion models through guidance. This can benefit creative applications such as digital art, design. However, enhanced quality may also increase the risk of misuse, including the creation of deceptive or harmful content like deepfakes. While our work does not involve identity synthesis or model release, we acknowledge this risk and recommend responsible deployment practices such as usage restrictions and content disclosure.

\clearpage
\begin{table}[!h]
\centering
\caption{\textbf{Prompt and seed list used for general quality improvement via HeadHunter.}}
\begin{tabularx}{\linewidth}{@{}Xc@{}}
\toprule
\textbf{Prompt} & \textbf{Seed} \\
\midrule
a close up of an old Japanese man, soft lighting, 8k & 0 \\
\hline
A black man with long braids, wearing white robes and heavy metal jewelry, pointing at the viewer against a dark background in the style of cinematic photography for a fashion shoot. & 0 \\
\hline
A photo of an Asian girl with her hand on her nose, with red nail polish and a bandaid, with black hair, a close up portrait, in the style of Rinko Kawauchi. & 0 \\
\hline
Japanese woman with short black hair, bangs hairstyle, hand covering eye, ring rings on, cool pose, outdoor sunlight background, natural lighting, portrait photography, in the style of Leica Q2 camera & 0 \\
\hline
A group of models of diverse ethnicities wearing various Nike sneakers, all dressed in white and colorful with patterns and designs inspired by the pop art movement, posed together for an editorial photoshoot against a backdrop featuring abstract collage-like elements in shades of red and blue. The scene was vibrant and dynamic, capturing their energetic expressions and stylish outfits in the style of the pop art movement. & 0 \\
\hline
photo of a black special forces soldier doing an olympic ski jump, holding two katanas in his hands, white background, 35mm film stills in the style of Cai Guo-Qiang and James & 0 \\
\hline
A tall German man sitting in a bench, reading a book, medium shot & 0 \\
\hline
A breakdancer doing a backflip & 0 \\
\hline
A photograph of an old oak tree in the middle, surrounded by dense woodland foliage, with sunlight filtering through the leaves and casting dappled light on the ground. The image was shot using a Hasselblad camera, Kodak Gold 200 film, and Portra 800 film stock. & 0 \\
\hline
a close-up of a woman's hand with delicate fingers, soft lighting, photorealistic & 0 \\
\hline
a photo of beautiful girl & 0-9 \\
\bottomrule
\end{tabularx}
\label{tab:general_prompt_list}
\end{table}

\begin{table}[!h]
\centering
\caption{\textbf{Content prompts and corresponding seeds used for style-oriented quality improvement via HeadHunter.}}
\begin{tabularx}{\linewidth}{@{}Xc@{}}
\toprule
\textbf{Prompt} & \textbf{Seed} \\
\midrule
portrait of an elderly man with white hair, wearing a wool coat, looking into the distance & 0 \\
\hline
young woman in a red dress, standing in the wind, eyes closed & 1 \\
\hline
teenage boy with messy hair, wearing headphones, sitting on a rooftop & 2 \\
\hline
smiling girl holding a cat, in a flower garden & 3 \\
\hline
a man leaning against a brick wall, hands in pockets, calmly observing the street & 4 \\
\bottomrule
\end{tabularx}
\label{tab:content_prompt_list}
\end{table}
\FloatBarrier
\end{document}